\documentclass[lettersize,journal]{IEEEtran}
\usepackage{amsmath,amsfonts}
\usepackage{array}
\usepackage[caption=false,font=normalsize,labelfont=sf,textfont=sf]{subfig}
\usepackage{textcomp}
\usepackage{url}
\usepackage{verbatim}
\usepackage{graphicx}
\usepackage{cite}
\usepackage{lscape}
\usepackage{xcolor}
\graphicspath{{./figs/}}
\usepackage{amssymb}
\usepackage{rotating,caption,booktabs}
\usepackage{enumerate}
\usepackage[pagebackref,breaklinks,colorlinks]{hyperref}
\usepackage{algpseudocode} 
\usepackage{multirow}
\usepackage[acronym]{glossaries}

\usepackage{makecell}
\usepackage{color,colortbl}
\usepackage[T1]{fontenc}

\begin{document}
\newacronym{mmw}{MMW-Radar}{Millimeter Wave Radar}
\newacronym{imu}{IMU}{Inertial Measurement Unit}
\newacronym{gps}{GPS}{Global Positioning System}
\newacronym{ads}{ADS}{Autonomous Driving System}
\newacronym{bev}{BEV}{Bird's-Eye-View}
\newacronym{mot}{MOT}{Mutiple Object Tracking}
\newacronym{hfov}{HFoV}{Horizontal Field of View}
\newacronym{vfov}{VFoV}{Vertical Field of View}
\newacronym{fps}{FPS}{fragment per second}
\newacronym{kf}{KF}{Kalman Filter}
\newacronym{ekf}{EKF}{Extended Kalman Filter}

\newcommand{\tocite}{\textcolor{blue}{~[cite]~}}
\newcommand{\eqpref}{Eq.~}
\newcommand{\secpref}{Sec.~}
\newcommand{\figpref}{Fig.~}

\newcommand{\eg}{e.g.,~}
\newcommand{\ie}{i.e.~}
\newcommand{\etal}{et~al.~}

\newcommand{\quotate}[1]{``#1''}

\title{A survey on deep learning approaches \\
for data integration in autonomous driving system}

\author{Xi Zhu$^*$, Likang Wang$^*$, Caifa Zhou, Xiya Cao, Yue Gong, Lei Chen$^\dag$
\thanks{Manuscript received DATE; revised DATE. }
\thanks{$^*$ denotes equal contribution.} 
\thanks{$^\dag$ Corresponding author. Email: leichen@cse.ust.hk}
\thanks{Xi Zhu, Caifa Zhou and Xiya Cao are with Riemann Laboratory, 2012 Laboratories, Huawei Technologies, China.}
\thanks{Likang Wang and Lei Chen are with Department of Computer Science and Engineering, Hong Kong University of Science and Technology, Hong Kong, China. This work is done when Likang Wang is an intern at Huawei Technologies.}
\thanks{Yue Gong is with Parallel Distributed Computing Laboratory, 2012 Laboratories, Huawei Technologies, China.}} 

\markboth{Journal Name,~Vol.~X, No.~X, Month~2023}%
{Shell \MakeLowercase{\textit{et al.}}: A Sample Article Using IEEEtran.cls for IEEE Journals}

\IEEEpubid{0000--0000/00\$00.00~\copyright~2021 IEEE}

\maketitle
\begin{abstract}

The perception module of self-driving vehicles relies on a multi-sensor system to understand its environment. Recent advancements in deep learning have led to the rapid development of approaches that integrate multi-sensory measurements to enhance perception capabilities. This paper surveys the latest deep learning integration techniques applied to the perception module in autonomous driving systems, categorizing integration approaches based on ``what, how, and when to integrate.'' A new taxonomy of integration is proposed, based on three dimensions: multi-view, multi-modality, and multi-frame. The integration operations and their pros and cons are summarized, providing new insights into the properties of an ``ideal'' data integration approach that can alleviate the limitations of existing methods. After reviewing hundreds of relevant papers, this survey concludes with a discussion of the key features of an optimal data integration approach.

\end{abstract}


\section{Introduction}
\label{sec:intro}


Perception is a crucial component of \acrfull{ads} \cite{janai2020computer, fayyad2020deep, chen2019autonomous}. It enables self-driving vehicles to perceive and comprehend their surroundings and accurately position themselves. The performance of the perception module significantly influences downstream tasks, such as planning and control, as well as driving safety. The two primary functions of the perception module are environment perception and localization \cite{fayyad2020deep}. Environment perception involves the vehicle actively gathering information about both static elements (e.g., lane lines, road markings, and traffic signs) and dynamic objects (e.g., other vehicles, pedestrians). In contrast, localization requires the vehicle to also consider its own motion state measurements, such as speed, heading, and acceleration. All these perception data can be obtained with the help of a sensor suite built-in the vehicle \cite{huang2022multi}. Although the sensor suite consists of a bunch of sensors \cite{fayyad2020deep, chen2020survey, kuutti2018survey}, the ones used for environmental perception are cameras, LiDARs, and \acrfull{mmw} \cite{janai2020computer, marti2019review, huang2022multi, ignatious2022overview}. Detailed properties of these sensors are further explained in Section~\ref{sec:sensor}. 

Although all sensors contribute to information collection, single-sensor systems have limitations and shortages that make it difficult to perform complete, accurate, and real-time environmental perception in autonomous driving applications \cite{kolar2020survey, geiger2012we, fayyad2020deep}. This is because different sensors have different temporal and spatial coverage, as well as areas of expertise and weakness \cite{wang2019multi, marti2019review, yeong2021sensor, morkar2022autonomous}. For example, cameras can capture high-resolution images with rich color information, but they cannot work well in low-visibility scenarios or provide reliable 3D geometry. Additionally, cameras' angular coverage is limited by their field-of-view. LiDARs are superior at 3D geometry estimation, have a wide range of view, and can work in dim light, but the points they capture are usually sparse due to the low sampling rate. Radars cannot acquire texture information, but they can capture the velocity of moving objects, which neither LiDARs nor cameras can do. Furthermore, single-sensor systems may suffer from the problem of deprivation, which describes the circumstances of perception loss or failure when the sensor stops working or cannot function well \cite{kolar2020survey}. High uncertainty and imprecision are also significant concerns of single-sensor systems when data are missing or or measurements are not accurate.
\IEEEpubidadjcol 

To address the limitations and challenges of single-sensor systems, various methods have been proposed to integrate data from different sensors in recent years \cite{kolar2020survey, feng2020deep, huang2022multi, wang2023multi}. In the literature of \acrfull{ads}, data integration is often referred to as data fusion, data integration, or sensor fusion, which are used interchangeably \cite{li2020multi, wang2019multi, yu2021real, meng2020survey}. Data integration involves logically or physically transforming information from different sensors or sources to obtain a more consistent, informative, accurate, and reliable output than what could be achieved by using one sensor alone \cite{bleiholder2009data, hall1997introduction, khaleghi2013multisensor, meng2020survey, velasco2020autonomous, yu2021real, zhou2019information, li2020multi, wang2019multi}. Data integration techniques can be broadly divided into classical algorithms \cite{kolar2020survey, campbell2018sensor, gruyer2017perception, van2018autonomous, fayyad2020deep, feng2021review} and deep learning approaches \cite{fayyad2020deep}, with the latter being the focus of this paper due to its increasing popularity and potential for higher accuracy results. Deep learning-based integration techniques have drawn increasing attention, thanks to the rapid development of deep learning in computer science and artificial intelligence \cite{wang2019multi, marti2019review, velasco2020autonomous, cui2021deep, yeong2021sensor, huang2022multi, feng2020deep, wang2023multi}. They not only yield better accuracy results but also eliminate the need for hand-crafted design and fully exploit information contained in big data per high computational power \cite{chen2020survey}. Data integration techniques have also been explored in other research fields, such as smart living \cite{cai2019survey, uddin2020body, zhang2017deeppositioning, liu2019design, wu2021multi}, medical field \cite{guo2019deep}, transportation \cite{shahrbabaki2018data, du2018hybrid, huang2018modeling}, industry \cite{guo2019multitask, torres2020multilevel, beard2020method}, and business \cite{guo2018mobile, ayata2018emotion, zhang2022credit}. However, since the current paper focuses on data integration in \acrshort{ads}, readers interested in other domains are referred to relevant reviews \cite{lau2019survey, nweke2019data, qi2020overview, muhammad2021comprehensive, pham2021fusion, alam2017data}. In the next, we will summarize the key issues that need to be addressed in data integration in ADS. \newline

\begin{figure*}[!t]
	\centering
	\subfloat[]{\includegraphics[width=2.6in]{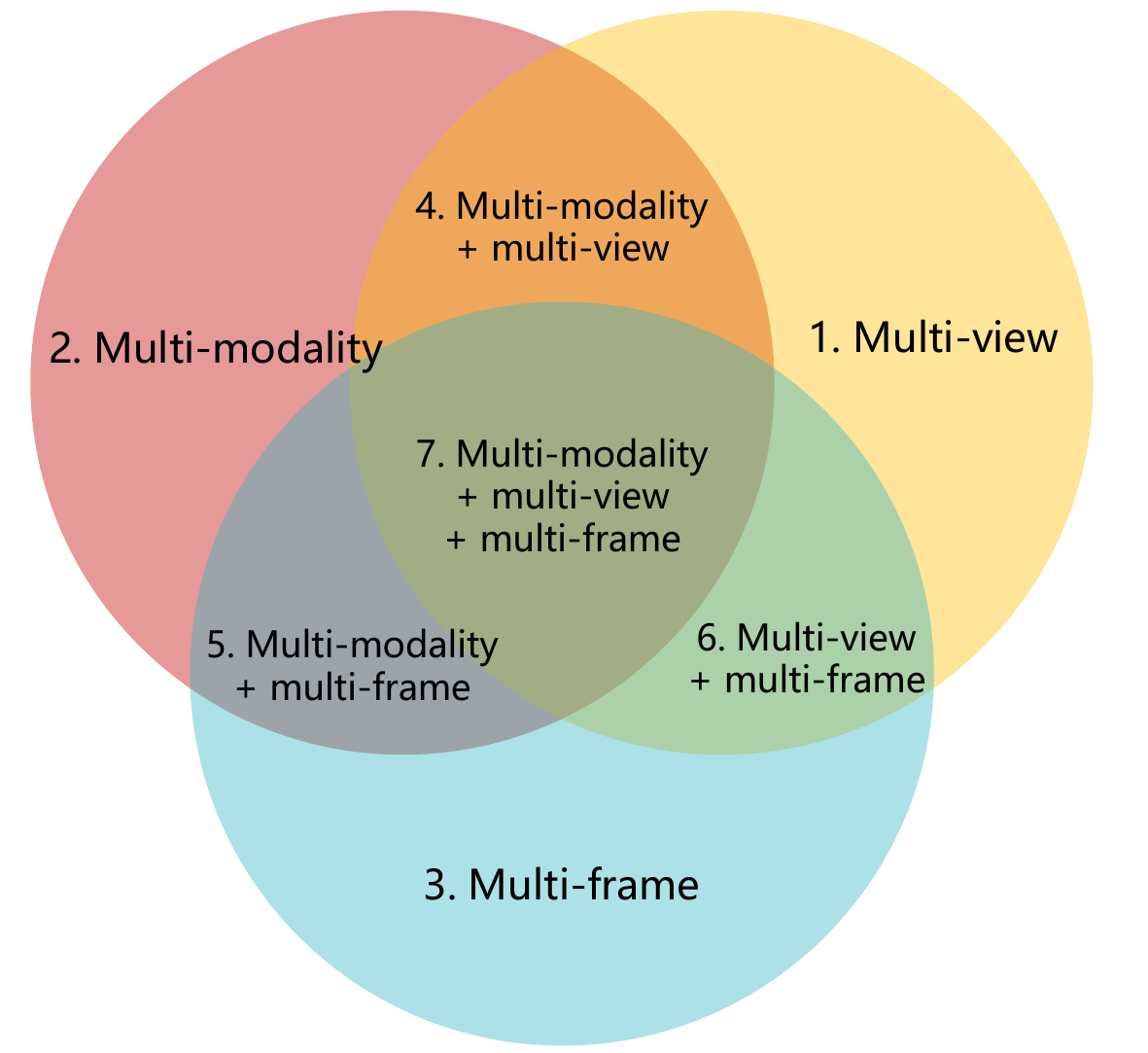}%
		\label{fig:bilevel_content}}
	\hfil
	\subfloat[]{\includegraphics[width=3.4in]{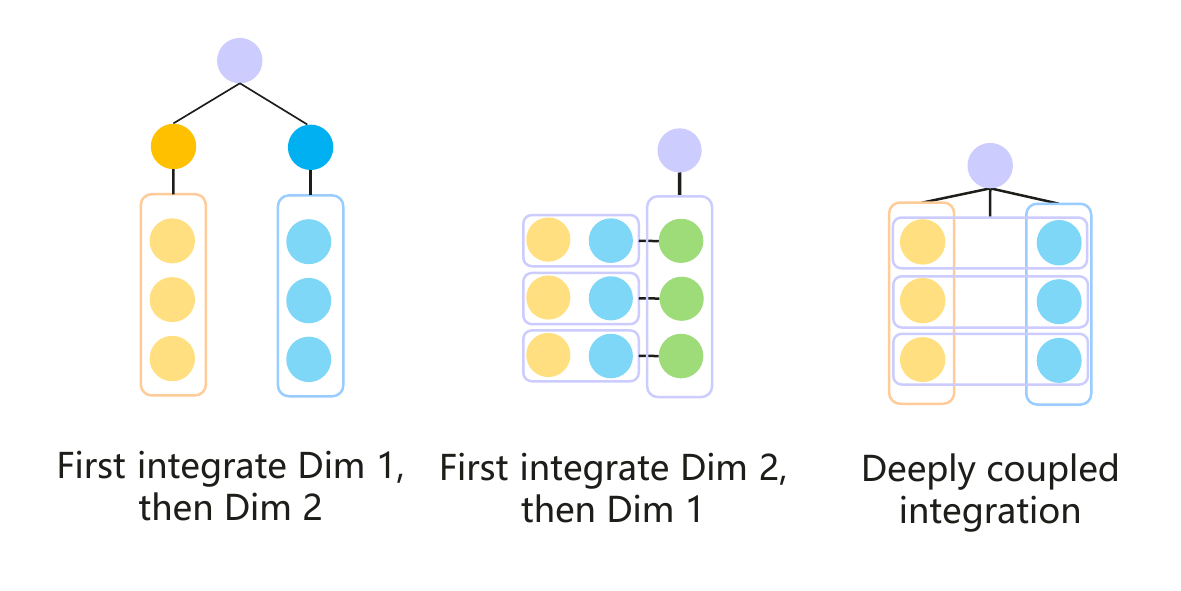}%
		\label{fig:bilevel_order}}
	\caption{Bi-level taxonomy of ``what to integrate''. (a) Seven upper-level categories based on three dimensions, describing the content of integration. (b) Three lower-level paradigms of each two-dimensional integration category, based on the order of integration.}
	\label{fig:bilevel}
\end{figure*}

\textbf{Key problems of data integration}

Three widely accepted key problems to be addressed in data integration include ``what to integrate'', ``how to integrate'', and ``when to integrate'' \cite{singh2019comprehensive, ravindran2020multi, feng2020deep}. Many reviews related to deep learning based data integration in \acrshort{ads} perception have been published in recent years, each of which partially covers these three problems and related techniques. While some reviews take data integration as a technical approach to address certain perception tasks under specific circumstances \cite{mo2022review, marti2019review, huang2020autonomous, gupta2021deep, zhou2020mmw, wang2022review, barbosa2023camera}, others center their attention on the data integration itself, including techniques, categories, and applications \cite{wang2019multi, fayyad2020deep, cui2021deep, yeong2021sensor, huang2022multi, feng2020deep, wang2021multi, wang2023multi}.
We mainly make comparisons with the latter.

\subsubsection{What to integrate}
\label{sec:intro-what}

``What to integrate'' is a question about the input to the integration process. In this survey, we divide ``what to integrate'' into two hierarchical sub-questions: what is the content to integrate, and what is the order to integrate. 

a) What is the content to integrate?

We divide the content of integration into three dimensions: 
\begin{itemize}

	\item[-] Multi-view integration involves integrating representations or views from multiple sensors of the same type that point in different directions. For example, integrating point clouds from multiple LiDARs placed at different locations on a car or integrating images taken from multiple cameras with different orientations \cite{zhou2020end}. However, combining different view representations of one LiDAR point cloud, such as integrating a \acrfull{bev} and a range view of LiDAR point cloud, is not considered multi-view integration since it does not introduce new information. These view transformations are considered part of the pre-processing of sensed data. 
	
	\item[-] Multi-modality integration involves integrating data from different types of sensors, such as cameras, LiDAR, and \acrfull{mmw}, in any combination. Research has shown that fusing camera and LiDAR data is one of the most common types of integration. 
	
	\item[-] Multi-frame integration involves integrating data over time, while the former two dimensions focus on spatial integration at one time step.
\end{itemize}

Given these three dimensions, we could classify all data integration methods into seven upper-level categories, as shown in Figure~\ref{fig:bilevel_content}. Data integration cannot only be implemented within one of three dimensions, but also be carried out over multiple dimensions. The former is denoted as single-dimensional integration.

b) What is the order to integrate?

Multidimensional integration, such as ``multi-modality and multi-view,'' is different from single-dimensional integration as it requires multiple combinations of operations. We further categorize multidimensional integration into lower-level integration paradigms based on the integration order, as illustrated in Figure~\ref{fig:bilevel_order}. For instance, in two-dimensional ``multi-modality and multi-view'' integration, there are three sub-level categories or paradigms: modality-first integration, view-first integration, and deeply coupled integration. Three-dimensional integration is even more complex, and to our knowledge, the corresponding paradigms have not been discussed in existing studies. Therefore, we limit our discussion of relevant literature to one- or two-dimensional integration in Section~\ref{sec:what}.

\subsubsection{When to integrate}
\label{sec:intro-when}

The topic of ``when to integrate'' pertains to the level of data abstraction at which integration occurs. This has been extensively covered in literature, with most adopting a widely used categorization scheme based on the input data abstraction stage. This scheme includes data-level (or pixel, signal, or low-level), feature-level (middle-level), and decision-level (or result or high-level) fusion \cite{fayyad2020deep, yeong2021sensor, feng2020deep, wang2021multi}. Data-level integration methods fuse raw or preprocessed data before feature extraction, while feature-level integration combines extracted features at intermediate neural network layers. Decision-level integration involves merging the output data separately estimated by each sensor. In addition to these levels of integration, a new category of ``when-to-integrate,'' called multi-level integration, has emerged in recent years as more studies attempt to integrate data at different levels \cite{cui2021deep}. An example of multi-level integration is using result-level 2D bounding boxes from RGB images to select data-level or feature-level 3D LiDAR data \cite{xu2018pointfusion, qi2018frustum, paigwar2021frustum}. Each level of data integration has its own advantages and disadvantages, and no evidence supports one level being superior to another \cite{feng2020deep}. In Section~\ref{sec:when}, we review existing integration methods based on these four categories and summarize their pros and concons.

\subsubsection{How to integrate}
\label{sec:intro-how}

The question of ``how to integrate'' pertains to the specific integration operations used to mathematically combine data. These operations are often overlooked or not explicitly stated since they are typically simple and straightforward, such as addition or concatenation. However, in our survey, we present several common integration operations, including projection, concatenation, addition/average mean/weighted summation, probabilistic method, rule-based transaction, temporal integration approaches, and neural network/encoder-decoder structure. We provide a summary of each integration operation's properties and practical applicationions. \newline

\textbf{Purposes and paper organization}

Our survey aims to offer a comprehensive overview of the latest (from 2017 to 2023) deep learning-based data integration techniques for camera, LiDAR, and \acrshort{mmw} in \acrshort{ads} perception, following the previously mentioned taxonomy. After reviewing hundreds of pertinent papers, we identify limitations in existing integration methods and present a discussion addressing the open question of the ideal data integration approach for \acrshort{ads} perception. 

\begin{table*}[t]
\centering
\begin{tabular}{ccccccccc}
\hline
Reference & Task & Sensor & Data representation  & Integration operation & Method Pros \& Cons & Dataset & Calibration & \\
\hline
 Wang \etal \cite{wang2019multi} & Various & $\pmb{\checkmark}$ &  $\pmb{\checkmark}$  &  &  &  &  & \\
 Fayyad \etal \cite{fayyad2020deep}  & Various & $\pmb{\checkmark}$ &   &  & $\pmb{\checkmark}$ &  & & \\
 Cui \etal \cite{cui2021deep} & Various &  &  $\pmb{\checkmark}$  &  & $\pmb{\checkmark}$ &  & $\pmb{\checkmark}$ & \\
 Yeong \etal \cite{yeong2021sensor}  & OD & $\pmb{\checkmark}$ & &  & $\pmb{\checkmark}$ &  & $\pmb{\checkmark}$ & \\
 Huang \etal \cite{huang2022multi} & Various &  &  $\pmb{\checkmark}$ &  &   & $\pmb{\checkmark}$  & &  \\
 Feng \etal \cite{feng2020deep} & OD \& SS &  $\pmb{\checkmark}$ & $\pmb{\checkmark}$  &  $\pmb{\checkmark}$ &  & $\pmb{\checkmark}$ & & \\
 Wang \etal \cite{wang2021multi} & OD & $\pmb{\checkmark}$ & $\pmb{\checkmark}$  &   & $\pmb{\checkmark}$ & $\pmb{\checkmark}$  & & \\
 Wang \etal \cite{wang2023multi} & OD & $\pmb{\checkmark}$ &   $\pmb{\checkmark}$ &  $\pmb{\checkmark}$ & $\pmb{\checkmark}$ &  $\pmb{\checkmark}$ &  & \\
 Ours & Various &  $\pmb{\checkmark}$ &   $\pmb{\checkmark}$ &  $\pmb{\checkmark}$ & $\pmb{\checkmark}$ &  &  & \\
\hline
\end{tabular}
\caption{Brief summary of recent reviews on data integration in ADS perception using deep learning approaches. ``OD'' represents object detestion, and ``SS'' means semantic segmentation.}
\label{tab:review-checks}
\end{table*}

In Table~\ref{tab:review-checks}, we briefly compare the content covered by related surveys on deep learning-based data integration with ours. The key contributions of our survey are summarized as follows:

\begin{itemize}
    \item We propose three fundamental dimensions (multi-view, multi-modality, multi-frame) for data integration, based on which a new bi-level taxonomy is created. The upper-level taxonomy outlines the content to integrate with seven categories, while the lower level examines the order of integration. This bi-level taxonomy allows for a comprehensive categorization of data integration approaches.
     
    \item We summarize the common integration operations used in deep learning models, as well as their pros and cons, which are often not fully explored in other reviews.
    
    \item Based on our review, we provide a detailed summary and analysis of the limitations of existing integration techniques in \acrshort{ads} perception and suggest our vision for the ideal data integration approach.
     
\end{itemize}

The remainder of this paper is organized as follows: In Section~\ref{sec:sensor}, we discuss the measuring principles, characteristics, advantages, and disadvantages of three types of sensors commonly used in \acrshort{ads} perception and their representations. We then offer an overview of ``what to integrate,'' ``when to integrate,'' and ``how to integrate'' in Section~\ref{sec:what}, Section~\ref{sec:when}, and Section~\ref{sec:how}, respectively. In Section~\ref{sec:casestudy}, we present a case study illustrating the practical applications of these three questions. In the final section, we summarize our work and discuss ideal data integration approaches and future directions.

\section{Sensing modalities and pre-processing}
\label{sec:sensor}
Commonly used sensors in \acrshort{ads} perception can be classified into two groups according to their operational principles \cite{yeong2021sensor, cui2021deep}: 
\begin{itemize}
	\item Exteroceptive sensors, such as cameras, LiDARs, and \acrshort{mmw}s, actively collect data of surroundings to perceive the external environment. 
	
	\item Proprioceptive sensors, including \acrshort{imu}, wheelmeter, \acrshort{gps} receiver, etc., are mainly used to capture the internal states of vehicles and the dynamic measurements of the system. From the perspective of tasks, these sensors are widely employed together with exteroceptive sensors for positioning and localization.
\end{itemize}

The data collected by the aforementioned sensors vary in terms of coordinate frames and characteristics, each with their own strengths and limitations. Integrating data from different coordinate frames can be challenging, and many existing deep learning architectures are designed to process specific data representations. To address this issue, a common approach is to apply pre-processing methods to transform raw data into appropriate representations.

In this section, we focus primarily on three types of exteroceptive sensors, discussing their properties and pre-processing methods (representations). Table~\ref{tab:sensors} provides a summary of the advantages and disadvantages of these sensors.

\begin{table*}[!h]
	\small
	\centering
	\caption{Comparison of camera, LiDAR, and radar sensors.}
	\label{tab:sensors}
	\begin{tabular}{cccccccccc}
		\hline
		Sensor & Data Format & Resolution & HFoV & Geometry & Texture & Bad weather & Dim/Dark & Velocity & Cost \\
		\hline
		Camera & 2D pixels & ++ & + & - & ++ & + & - & - & Low\\
		
		LiDAR & 3D points & + & ++ & ++ & - & + & ++ & - & High \\
		
		Radar & 3D points & + & + & + & - & ++ & ++ & ++ & Low \\
		\hline
	\end{tabular}\\
	\vspace{1ex}
	++: Comparatively has strong capability. +: Has limited capability. -: Comparatively has weak capability.
\end{table*}

\subsection{Camera} 
\label{sec:sensor-camera}
Cameras are optical devices capable of capturing 2D visual images. While some cameras (\eg infrared) can detect invisible light, the term 		``\textit{camera}'' usually refers to those sensing visible light. Cameras can generate both grayscale and colored images, although most modern cameras default to producing colored images. Some cameras (\eg gated, time of flight (TOF), structured light) emit waves and detect their responses, but we do not focus on these as they operate similarly to LiDAR. Instead, this section concentrates on RGB cameras, the most common optical sensors, which passively receive visible light with wavelengths between 400 and 700 nm and output colored images.


In general, cameras can be modeled with a pinhole model. Each point in 3D space is projected to a pixel according to an affine transformation determined by the projection matrix. This projection is related to both the camera's intrinsic properties and its pose in the 3D world. They can be described by two affine transformations determined by the intrinsic and extrinsic matrices, respectively. Most cameras have only one lens, called monocular, while others, called stereo, may have multiple lenses. Stereo cameras mimic human binocular vision and can perceive 3D objects with algorithms. Monocular cameras can also recover 3D information by filming the same object with multiple cameras or from multiple poses. The main difference between these two types is that the relative positions of lenses and directions are fixed and known inside a stereo camera but are unknown and require further estimation for monocular systems.



RGB cameras aim to reproduce images perceived by human eyes and are essential for capturing colored and textured regions (\eg road lanes, traffic signs, and traffic lights). These cameras generally have very high spatial (hundreds to thousands in height and width) and temporal (dozens to hundreds of frames per second) resolutions. They often have a long working range, with the maximum perception distance reaching 1 km in good weather conditions. Their ability to capture color and texture information aids semantic comprehension. Moreover, cameras have low energy and manufacturing costs, allowing widespread deployment. However, RGB cameras have limitations in lighting conditions and line-of-sight visibility. Due to their working principle, they struggle to detect lightless objects (\eg road lanes and obstacles) in poorly lit scenes and have weak occluded object detection capabilities. Additionally, single-frame images taken by monocular cameras lack geometric information, requiring multiple images and complex algorithms (\eg depth recovery and 3D reconstruction) for depth and 3D structure estimation. Cameras are also vulnerable to external environments, as the lens may be blurred by liquid (\eg rain, spray, and wheel splash), and distant objects may not be recognizable in fog.

In ADS, the integrated camera data are represented in either 2D or 3D formats.

\subsubsection{Pixel representation}
The representation of pixels involves storing pixel features on a 2D image plane, where each pixel has multiple channels to describe its properties. This results in the entire image being stored in a 3D matrix with dimensions of $[height,\ width,\ channel]$. Typically, RGB raw images have three colored channels, but other cameras may have different channels such as depth, gray, infrared, or gated channels.

\subsubsection{Point or voxel representation}
The representation of points or voxels takes into account the depth information by projecting each pixel into 3D space. These 3D points can be stored as either point clouds or voxel grids. Point clouds assign each point a float number 3D coordinate, resulting in a matrix with dimensions of $[n,\ c+3]$, where $n$ represents the number of pixels and $c$ represents the number of channels. Voxel grids divide the space into grids with dimensions of $[height,\ width,\ depth]$, and points are placed into these grids.

\subsection{LiDAR}
\label{sec:sensor-lidar}

Light Detection and Ranging, a.k.a. LiDAR, is a technique commonly used for range measurements in autonomous driving \cite{li2020deep}. Its working principle is to estimate the time intervals between emitted light pulses and received signals reflected by target objects, and obtain the distances with the time intervals and light speed. 

Three types of LiDAR, including 1D, 2D, and 3D LiDAR, are used to collect different amount of environment information \cite{yeong2021sensor}. While 1D LiDAR can only provide measurement of distance, 2D LiDAR can obtain the spatial information on an X-Y coordinate horizontal plane of the target by spinning certain degrees horizontally. The horizontal degree a LiDAR sensor rotates over is called the \acrfull{hfov} of the sensor. 3D LiDAR sensors expand the vertical view by firing multiple lasers vertically, making the data collected in the 3D X-Y-Z coordinate system. 3D LiDAR sensors are more commonly employed in autonomous vehicles, while the high price is the concern in implementation \cite{wang2019multi, qian2020end}. 

The process of generating data using LiDAR sensors involves the use of beams of light to draw samples from the surfaces of objects in the surrounding environment. This laser-firing working principle allows LiDAR sensors to function well in low-visibility conditions but makes them susceptible to external weather conditions such as rain, fog, snow, and dusty environments \cite{kutila2018automotive}. Additionally, the color of the target can impact the performance of LiDAR sensors, with darker-colored objects absorbing light and being less reflective than lighter-colored ones \cite{petrovskaya2008model}.
The sampling range of the scene is determined by the \acrshort{hfov} and \acrfull{vfov}, while other parameters such as horizontal/vertical resolution and \acrfull{fps} contribute to the data intensity. Horizontal and vertical resolution refer to the density of sampling in space, with smaller resolution resulting in denser sampling given fixed \acrshort{hfov} and \acrshort{vfov}. \acrshort{fps} describes the sampling density in time, i.e., how many scans the LiDAR conducts per second. Detailed specifications of different LiDAR sensors from different companies are provided by Yeong \etal \cite{yeong2021sensor}.

Unlike camera images, 3D LiDAR measurements are a set of irregular and unordered data points, referred to as point clouds in 3D structure \cite{cui2021deep}. To fit the input format of different deep learning models, point clouds can be transformed into several different representations using pre-processing methods. It is worth noting that LiDAR data is sparser compared to image data.

\subsubsection{Point representation}

3D point clouds obtained from LiDAR sensors can be processed without format transformation with point processing deep learning networks such as PointNet \cite{qi2017pointnet}, PointNet++ \cite{qi2017pointnet++}, PointCNN \cite{li2018pointcnn}, and KPConv \cite{thomas2019kpconv}. LiDAR point clouds can be integrated with similar point-format data such as other LiDAR point clouds \cite{wu2021pedestrian}. Though point clouds retain the original information and may provide larger receptive field \cite{qi2017pointnet++}, the volume of point clouds can be huge, requiring high computation power to process \cite{wang2021multi, cui2021deep}. Moreover, it is hard to integrate point clouds with other data formats such as images. Due to these two points, representations with additional pre-processing methods are developed and progressed rapidly.  

\subsubsection{Voxel representation}

Voxels are generated by dividing the whole 3D space into small regular 3D grids and partitioning the original points into corresponding grid based on the geometry. This gridization transforms irregular points into regular voxel representation, and make it possible to down-sample the original LiDAR points to reduce input volume. In fact, the volume and the resolution of the voxels can be adjusted by changing the grid size. Larger grids result in more information loss while smaller grids may still bring burdens to computation. Several 3D convolution methods can be used to process voxels and extract features, such as 3D ShapeNet \cite{wu20153d}, VoxelNet \cite{zhou2018voxelnet}, and VoxNet \cite{maturana2015voxnet}.

\subsubsection{Pixel/View representation}

Pixel or view representation of LiDAR points is to convert 3D point clouds into 2D image views by projection. \acrfull{bev} and range views (also known as perspective views) are two common types of views on different 2D view planes that can be transformed from point clouds to. Pixel representation can leverage the existing well-developed CNN-family image processing methods, though the 3D geometry information retained in original point clouds may be lost in the projection process. The pixel representation, or projection technique, is commonly adopted when integrating LiDAR point clouds with camera images \cite{sobh2018end, bai2022transfusion, wang2019latte}.

\subsubsection{Integrated representation}

In addition to the previously mentioned representations, some researchers have attempted to combine various representations to obtain richer information or before further processing. They have developed point-voxel integration methods that merge the advantages of both representations. Point-Voxel CNN (PVCNN) \cite{liu2019point} employs a dual-branch system consisting of a voxel-based branch and a point-based branch. It carries out convolutions on voxels while supplementing detailed geometry information from points. PV-RCNN \cite{shi2020pv} generates 3D proposals from voxel convolutions and selects a few key points in the space to serve as connections between voxel features and the refinement network for these proposals. Furthermore, different views obtained from point clouds can be integrated. Zhou \etal \cite{zhou2020end} propose a Multi-View Fusion (MVF) Network that converts \acrfull{bev} and perspective views into voxels, fusing them together so that the complementary information can be used effectively. There are also studies integrating all aforementioned representations. In M3DETR \cite{Guan_2022_WACV}, the point, voxel, and pixel representations are processed with PointNets, VoxelNet, and 2D ConvNets modules respectively, and fused with transformers at multiple scales. This design better extracts information in raw data by effectively exploiting the correlation between different representations leveraging the attention structure.

\subsection{Millimeter wave radar (MMW-radar)}
\label{sec:sensor-radar}

Radio Detection and Ranging, a.k.a. radar or \acrshort{mmw}, is a technique that relies on radiating electromagnetic millimeter waves and scattered reflections to estimate the range information about targets \cite{yeong2021sensor}. Short-range, medium-range, and long-range radar detectors have different detection distances, and are commonly used for collision avoidance and obstacle detection. 

Different from LiDAR sensors and cameras that are easily affected by external conditions, radar sensors are more robust in extreme weather or dim light \cite{zhang2018real, wei2022mmwave, zhou2020mmw}. Another significant advantage of this type of low-price sensor is its capability of accurately detecting the velocity of dynamic targets based on the Doppler effect, which is very important for perception tasks in autonomous driving scenarios. However, radars also have some disadvantages. Comparing with cameras, radars lack texture or semantic information. Comparing with LiDAR sensors, radars have lower angle resolution (see Table 4 in \cite{yeong2021sensor} for detailed configurations). Therefore, radars are not suitable for tasks such as object recognition, and may have troubles when distinguish static and stationary objects \cite{wang2019multi, yeong2021sensor}. Besides, the clutters, i.e., the unwanted echoes in electronic systems, may cause false detections and performance issues in radar systems \cite{zhou2020mmw}. 

According to Wang \etal \cite{wang2021multi} and Zhou \etal \cite{zhou2020mmw}, the data format of radar can be divided into raw data, cluster-layer data, and object-layer data according to different pre-processing stages. The raw output of radar is in the form of time frequency spectrograms. To improve its utility, signal processing methods such as those detailed presented in Zhou \etal \cite{zhou2022towards} are often necessary. More commonly adopted radar data formats in autonomous driving applications are cluster-layer obtained after operating clustering algorithms, and object-layer after filtering and tracking. Comparing with the original raw data, the latter two formats provide more sparse and less noisy information. 

Two different representations of radar signals can be found in \acrshort{ads} related research. 
\subsubsection{Point representation}
Point-based representation is to represent and process radar data as point clouds \cite{schumann2018semantic}. However, as raised by Wang \etal \cite{wang2021multi}, the properties of radar point clouds differ from LiDAR point clouds, thus issues may arise when directly using LiDAR models on radar points. 
	
\subsubsection{Map representation}
Another representation is map-based, which is to accumulate radar data over several time-stamps and generate radar grid \acrshort{bev} maps \cite{feng2020deep}. Since grid maps alleviate the radar data sparsity issue, image processing networks such as CNN are used to extract features and get static environment classification. More details of different grid maps can be found in \cite{zhou2020mmw}.

\section{Data Integration: What to Integrate}
\label{sec:what}
In this section, we review and summarize the techniques with respect to the content to integrate by following the bi-level taxonomy presented in Figure~\ref{fig:bilevel}. Specifically, we first summarize each one-dimensional integration as shown in Zone 1, 2, and 3 in Figure~\ref{fig:bilevel_content}. We then provide discussions for two-dimensional categories (Zone 4, 5, 6 in Figure~\ref{fig:bilevel_content}), with an additional focus on the order to integrate (Figure~\ref{fig:bilevel_order}). As research related to three-dimensional integration (Zone 7 in Figure~\ref{fig:bilevel_content}) is rare, we omit three-dimensional integration in this section. 

\subsection{Multi-view integration}
\label{sec:what-multiview}

Under our definition, the most common type of multi-view data integration is to integrate multi-view camera images captured by multiple cameras from different directions (e.g., 6 monocular cameras installed on self-driving vehicles). This camera multi-view integration can be used to generate \acrshort{bev} representation of the surrounding environment \cite{pan2020cross, li2021hdmapnet, zhou2022cross} or assist object detection in 2D images \cite{wang2022detr3d}. A stereo camera can also generate multi-view images as it has two or more lens \cite{ramos2017detecting}. Besides visual sensors, researchers also explore alternative methods to fuse multiple LiDARs' measurements mounted on a vehicle for object detection \cite{wu2021pedestrian, cao2020obstacle}. Due to high expense of LiDARs, however, this approach is not widely adopted in industry and relevant works are limited. Thus we skip the discussion of Lidar multi-view integration. Similarly, multi-radar integration is also considered as helpful in 360-degree environmental perception \cite{roos2019radar, dickmann2016automotive}, while they are more commonly integrated with other sensor modalities in applications.

\subsubsection{Camera multi-view} 

Most monocular RGB cameras can only capture objects within a frustum whose shape is determined by camera lens. Besides, many objects inside the frustum cannot be observed in images if occluded by others. Thus, we need to have multiple images shot from different positions to describe the 3D world more thoroughly. The depth information (distance between a point in 3D space and the camera's image plane) can be recovered from multi-view cameras given their relative camera parameters (intrinsic and extrinsic transformation matrix) based on the consistency and uniqueness of spatial geometry. Intrinsic parameters describe the mapping from 2D camera coordinate to 2D image coordinate, and extrinsic ones describe the mapping from 3D world coordinate to 2D camera coordinate. If images are taken from two paired cameras, it is called binocular depth estimation or stereo matching, or disparity estimation. If more than two views are provided, it is called multi-view stereo (MVS). 

The inputs of most camera multi-view methods \cite{oliveira2020topometric, ramos2017detecting, ismvsnet, song2023prior, yu20233d, lei2022c2fnet, sun2022multi, tian2022accurate, zhu2022garnet, peng2023bevsegformer, jiang2022polar, tao2023weakly} are RGB images, but there are also papers \cite{lee2021adversarially, vachmanus2021multi, sun2020fuseseg} fusing RGB and thermal images, and papers \cite{pan2020cross} fusing RGB and depths. We categorize camera multi-view inputs as following:

a) Images: These images are captured by different cameras or by a monocular camera but from different positions and angles. 
The number of images is not limited, but at least two images are required. The images should be overlapped but not totally the same.

b) Depth images: A 2D image with one channel, where each pixel denotes the camera's distance to the corresponding 3D point. They can be obtained from either depth cameras or depth estimation algorithms.

c) Thermal images: A 2D image with one channel, where each pixel denotes the intensity of infrared radiation emitted by an object. They can be captured by thermal cameras.

\subsubsection{Radar multi-view}

Very few papers investigate integration among multiple radars. \cite{ouaknine2021multi} uses three kinds of views: range-Doppler (RD), angle-Doppler (AD), and range-angle (RA). The RD view reveals the distance and velocity of objects. Specifically, non-moving objects respond at zero Doppler when the radar is stationary, and objects moving relative to the radar respond at nonzero Doppler. The AD view shows the direction and speed of objects. RA view represents the relationship between range and angle. All the views can be represented as 2D images.

\subsection{Multi-modality integration}
\label{sec:what-multimodality}

Since sensors have their own advantages and shortcomings as shown in Section~\ref{sec:sensor}, multi-modality data integration is expected to utilize the mutual supplementary between different sensors and achieve improved accuracy \cite{hall1997introduction}. Various methods have been developed and applied to tasks including object detection and tracking, depth completion, and segmentation. A few studies employ camera-radar integration \cite{wu2009collision, jha2019object, kowol2020yodar, liu2021robust, nabati2021centerfusion} or LiDAR-radar integration \cite{qian2021robust} for object detection. Several research further integrate all these three sensors together for environmental perception \cite{cai2020probabilistic, shahian2019real}. 

In the following, we summarize the input of existing multi-modality integration research for four multi-modality combinations: 1) camera and LiDAR, 2) camera and radar, 3) LiDAR and radar, and 4) camera, LiDAR, and radar. As for the remained combination of modalities, there is few research on them.

\subsubsection{Camera and LiDAR integration}

Since 2D camera images and 3D LiDAR points are in different coordinate frame, they usually need to be transformed to the same space before integration. In the following, we present discussions of 2D space integration and 3D space integration, respectively.

a) 2D space integration

Two sub-approaches can be identified according to different LiDAR projected planes. One approach is to integrate LiDAR and camera views in the same plane \cite{li2022deepfusion}. In other words, the LiDAR points are projected onto the image plane for integration. For example, Caltagirone \etal \cite{caltagirone2019lidar} transforms LiDAR points to dense LiDAR images by projecting LiDAR points to image plane and up-sampling. The LiDAR images then can be integrated with RGB images for road segmentation. Berrio \etal \cite{berrio2021camera} projects LiDAR points to image with mask techniques and probabilistic distribution to handle occlusions in semantic mapping. Object detection model, EPNet \cite{huang2020epnet}, projects LiDAR points to image plane for multiple times in order to enhance the LiDAR point features with corresponding image semantic information. Cheng \etal \cite{cheng2019noise} transforms LiDAR points to right and left lens of stereo, respectively, and integrate across both modality and lens to get dense depth. Models of \cite{hu2021penet, yan2021rignet, fu2019lidar} integrate RGB images with sparse depth map generated from LiDAR points for image depth completion. \cite{vora2020pointpainting, geng2020deep, liu2020sensor, ouyang2018multiview, pfeuffer2019robust, mendez2021camera, meyer2019sensor} also take RGB and LiDAR-generated depth image as inputs and return classification, bounding boxes, or mask proposals. There is also a study \cite{kim2022camera} expresses and fuses LiDAR features in the same coordinate system as the camera. \cite{zhang2022ri} converts LiDAR points to range images of size $(5, w, h)$ before integration with RGB images. 

Another approach is to integrate views in different planes. For example, RGB image and \acrshort{bev} map generated by LiDAR projection are integrated in \cite{bai2022transfusion, liang2018deep, schroder2019feature} with deep learning architectures. These works connect image feature map to \acrshort{bev} with different transformation or association methods. In \cite{zhao2018object}, four LiDAR projection maps, including height bird view (HBV), intensity bird view (IBV), distance center view (DCV), and intensity center view (ICV), are integrated with RGB images and original LiDAR 3D points.

b) 3D space integration

One way to achieve 3D space integration is by projecting image data to 3D space and thus generate \quotate{pseudo-LiDAR points}. As RGB images lack depth estimates, a significant portion of these works leverage stereo or RGB-D data to obtain accurate geometry information. \cite{wu2022sparse, liang2019multi} integrate LiDAR point cloud and pseudo-LiDAR point cloud generated from image depth completion for 3D object detection. Another way to integrate LiDAR and camera data in 3D space is via data association. 3D object detection models such as \cite{paigwar2021frustum, qi2018frustum, rovid2019towards} project 2D bounding boxes obtained in RGB image to 3D space to get frustums as regions of interest to guide 3D point or feature searching. VPFNet \cite{zhu2021vpfnet} leverages virtual points from 3D proposals as bridge to associate image features and aggregate LiDAR point features. In LoGoNet \cite{li2023logonet}, image features are associated with LiDAR generated 3D voxel features by projecting voxel point centroid using camera projection matrix and generate reference points in the image plane.

\subsubsection{Camera and radar integration}

Similar to LiDAR-RGB integration, camera and radar can also be integrated in either 2D space or 3D space. Examples of 2D space integration include \cite{kowol2020yodar, dong2021radar}, in which radar points are projected onto the 2D plane of camera's perspective view. In \cite{qi2022millimeter}, two-channel (radar cross section and range channels) radar image is generated with the same size as the visual image and combined with image feature maps with an extended VGG network. In 3D space integration, image information need to be converted to 3D space. RGB image in \cite{li2019vehicle} is first processed by CNN for feature extraction, and then converted to 3D points via back-projection. Radar and RGB can also be combined with data association. For example, CenterFusion \cite{nabati2021centerfusion} projects 2D object bounding boxes to 3D space to connect image information with 3D pillars generated from radar points.

\subsubsection{LiDAR and radar integration}

Since both LiDAR and radar data capture 3D geometry information by nature, they are usually integrated in 3D space. Some researches integrate raw data of the two sensors in a point-by-point way. For example, \cite{fritsche2017modeling} combines LiDAR and radar points based on heuristic rules to remove low-quality data affected by external environmental factors. LiDAR and radar points can also be combined via data association. \cite{kwon2016low} relies on radar detection to obtain frustums as regions of interest to filter relevant LiDAR points. 

\subsubsection{Camera, LiDAR, and radar integration}

Few studies are trying to integrate data from three sensors together for perception. The key process of the integration is still to transform the data from different sensors into the same space. Camera, LiDAR, and radar can be integrated in 2D space if LiDAR and radar points are projected to camera's plane. The input of the integration neural network in \cite{ravindran2022camera} is camera image layers concatenated with projected LiDAR and radar channels. The sensors can also be integrated in higher dimensional space. \cite{wang2020high} integrate RGB images, LiDAR points, and radar points by aggregating data information from each sensor together to generate high dimensional points before further processing. 

\subsection{Multi-frame integration}
\label{sec:what-multiframe}
Multi-frame data, i.e., temporal data, refer to the data sampled at multiple timestamps along timeline. Compared to single-frame data, integration of multi-frame data brings more information, and can potentially improve the accuracy of environmental perception. In this subsection, we focus on the inputs of single-sensor multi-frame integration. As few studies can be found on radar temporal integration, we only present details on camera image sequence and LiDAR point cloud sequence integration.

\subsubsection{Camera image sequence}

Camera multi-frame integration refers to fusing information of image sequences instead of a single image. Two types of data are commonly used to carry out the integration. One is the sequence of feature maps generated from each image, and the other is the sequence of processed information obtained from each pair of consecutive images.  

a) Sequence of feature maps generated from each image

A traditional two-stage temporal integration approach is to perform feature extraction operations corresponding to different tasks in each image frame and then create association or fusion cross frames. For example, in 2D \acrshort{mot} models such as \cite{cao2022observation, xu2019spatial, voigtlaender2019mots, liu2022opening, zhang2022multiple, sun2022multi, tian2022accurate}, 2D object detection bounding boxes and feature vectors/maps are first generated in each image, and then associated in different frames. Zhou \etal \cite{zhou2020tracking} conduct tracking in a similar way, yet they introduce the center of feature map for each image as an additional input. The same two-stage pattern can also be observed in 3D monocular \acrshort{mot}, while geometry or depth information is additionally incorporated \cite{osep2017combined, nguyen20203d}. Multi-frame integration of visual odometry (VO) algorithms also decouple feature extraction and temporal integration. For example, \cite{wang2017deepvo, xue2019beyond} use CNN to extract features and RNN to integrate information from different frames. Monocular depth estimation also takes RGB image sequence as inputs to extract information and integrate over time \cite{cs2018depthnet, patil2020don, zhang2019exploiting}. 

One disadvantage of this two-stage approach is that the results of the later stage may be affected by the earlier stage. To avoid this issue, some studies leveraging Transformer to integrate both spatial and temporal information for multi-frame image \acrshort{mot} have emerged in recent years \cite{sun2020transtrack, zeng2021motr, meinhardt2022trackformer}. 

b) Image pairs or sequence of processed information obtained from image pairs

Raw image pairs, specifically a preceding image and a current image, can directly be input into integration algorithms. For example, monocular depth estimation methods \cite{watson2021temporal, godard2019digging, zhou2017unsupervised, yin2018geonet, flora} utilize reference and target frame pairs for self-supervision. In these methods, features of the preceding image are re-projected to the view of target image with predicted depth and pose, thus a re-projection loss can be generated. In \cite{tao2023weakly}, image pairs from adjacent frames or different cameras can both be treated are images taken from different viewpoints with relative positions, and thus be used for weak supervision. Successive image pairs can also be exploited in knowledge distillation. By utilizing a three-level knowledge distillation method, Chen \etal \cite{chen2023real} train the student model to incorporate adjacent frames which allows it to acquire a greater understanding of comprehensive representation knowledge from its corresponding teacher model. 

In addition, the outputs of multiple image pairs can further be fused temporally. In other words, this approach consists of two stages. First, raw image pairs are combined respectively to get corresponding fused outputs. Then, the fused output sequence are integrated over time. One typical example of this approach is to generate optical flow from each pair of raw images and then integrate sequence of optical flows to, for instance, get or assist pose estimation \cite{zhou2018deeptam, li2019sequential, wang2020tartanvo, xue2019beyond, zhang2022multiple}.

\subsubsection{LiDAR scan sequence}

LiDAR multi-frame integration refers to the fusion of LiDAR scan sequence. As LiDAR scans have different representations, two types of LiDAR scan sequence input can be found in existing works: point cloud sequence, and view sequence.

a) Point cloud sequence

Point cloud sequence refers to multiple 3D LiDAR point clouds measured in continuous time series. Similar to image sequence integration, two-stage approaches can be applied here: first to extract features or obtain target objects from each 3D point cloud, and then to create association between LiDAR scans. In \cite{yin2021center, wang2022lidar, weng20203d, chiu2020probabilistic, chen2022mppnet, gao2023spatio, li2023modar}, the inputs of temporal integration are the sequences of extracted features or other forms of outcomes generated from LiDAR point clouds to achieve 3D object tracking. Another approach is to generate a denser point cloud by combining multiple LiDAR scans in order to enhance the data quality. This is usually done via data alignment, which means to align multiple point clouds to a unified coordinate frame. Wang \etal the conduct the cross-frame integration by aligning and fusing \quotate{thing class} data points from multiple consecutive LiDAR scans to create an enriched point cloud \cite{wang2022lidarseg}. The enriched point cloud is then treated as single-frame input for later processing stages such as sampling, training, and segmentation.

b) View sequence

Another type of input for LiDAR temporal integration is view sequence, which is a sequence of 2D view representations generated from 3D point clouds via projection. Since this approach convert 3D data sequence to 2D format, image-based temporal integration approaches can be leveraged for processing. \cite{duerr2020lidar} projects 3D points scans to range view images for semantic segmentation. The view image sequences (specifically, the feature maps extracted from views) are served as the input of temporal integration. LO-SLAM \cite{li2019net} converts LiDAR 3D point clouds to cylinder views with cylinder projection. This model takes view pairs generated from pair-wise scans as input to infer relative 6-DoF pose.

\subsection{Multi-view multi-modality integration}
\label{sec:what-multiview-multimodality}
Researchers explore possible approaches to introduce multiple views into camera and LiDAR/radar integration to enhance the perception capability. Several works, including~\cite{kummerle2018automatic, xiao2020multimodal, cheng2019noise, dinesh2020stereo, zhao2018object, natan2022towards, liu2022bevfusion}, integrate LiDAR with multiple camera images from different views. Some others such as \cite{zhao2018object, wu2021way, wang2020multi} integrate multi-view LiDARs and monocular camera. MVFusion \cite{wu2023mvfusion} propose a multi-view radar and multi-view camera fusion approach for 3D object detection. A mainstream pattern can be observed from these studies: first intra-modality (to fuse multiple views of a single modality), and then inter-modality (to fuse information from different modalities). Besides, research is also being conducted where multi-modality fusion is carried out first, followed by multi-view combination. In \cite{lou2023slam}, LiDAR points are integrated with monocular camera to obtain single frame 3D point cloud at each view. Then multiple 3D point clouds obtained are incrementally spliced for excessive 3D point cloud reconstruction with global optimizations. 
 
There is also research that fuses all three modalities: camera, LiDAR, and radar. For example, in \cite{chen2022futr3d}, the raw LiDAR inputs are in the form of point clouds, and features to be integrated are in the form of a multi-scale \acrshort{bev}. As for radar modality, the raw inputs are properties of radar points (such as location, speed, and intensity), and the features to be integrated are per-point hidden representations extracted by MLPs. Besides radar and LiDAR data, inputs also contain images from different views and corresponding camera parameters.

\subsection{Multi-view multi-frame integration}
\label{sec:what-multiview-multiframe}
Multi-view multi-frame integration introduces temporal relation to multi-view fusion process. Most papers focus on adding time-series information into camera multi-view integration, and few works investigate LiDAR or radar multi-view. Therefore, we only present a brief discussion on camera multi-view multi-frame integration algorithms in this subsection. 

\acrshort{bev} is an appropriate representation for multi-view camera data, thus some multi-view multi-frame papers adopt it into the integration pipeline. In \cite{li2022bevformer, can2020understanding, xiong2023neural, zhu2023nemo, liu2022petrv2, huang2022bevdet4d, zhang2022beverse, hu2021fiery, wang2023frustumformer}, images from different views are fused into a \acrshort{bev} representation, further propagating information along the temporal dimension for perception and prediction. It can be observed that in general, these algorithms tend to prioritize the step of spatial integration (multi-view) over temporal integration (multi-frame) in the fusion process.

Besides \acrshort{bev}, many works use regular image representations for temporal view fusion. \cite{li2018undeepvo} and \cite{li2020joint} use view pair sequences from stereo cameras to recover object scale or track objects. As stereo cameras are not always available, a great number of papers \cite{DBLP:conf/iros/JiaPCCLY20, wang2020tartanvo, li2019net, chen2019selective, clark2017vinet, xue2019beyond, DBLP:conf/cvpr/ZhanGWLA018, DBLP:conf/cvpr/DuzcekerGVSDP21, DBLP:conf/cvpr/SunXCZB21, stier2021vortx, wang2018mvdepthnet, long2020occlusion, long2021multi, rich20213dvnet, liu2022petrv2, han2023} utilize neighboring frames to form image pairs. These papers mainly target SLAM and reconstruction, which require multiple views to recover geometry. 

\subsection{Multi-modality multi-frame integration}
\label{sec:what-multimodality-multiframe}

Multi-modality multi-frame integration introduces a temporal aspect to the spatial integration process for enhanced perception performance. Current research on camera and LiDAR sequence integration is primarily focused on \acrfull{mot} and 3D object detection. For instance, 3D \acrshort{mot} models combine features or detected objects from LiDAR and camera to obtain instances in a single frame, followed by a data association process for multi-frame tracking \cite{kim2021eagermot, weng2020gnn3dmot, frossard2018end, wang2023camo}. The Transformer-based 3D object detection model proposed by Zeng \etal \cite{zeng2022lift} generates gridwise \acrshort{bev} images with RGB and LiDAR points, and then applies multi-sensor temporal integration operations given the \acrshort{bev} grids. These integration algorithms typically follow a two-stage paradigm: first, spatial integration (multi-modality fusion in a single frame), and then temporal integration (multi-frame association) \cite{shenoi2020jrmot, zhang2019robust, frossard2018end}.

There is limited research on the integration of LiDAR and radar point sequences. In one study \cite{qian2021robust}, 2D regions of interest are detected in the \acrshort{bev}s converted from LiDAR and radar 3D points. The features are first integrated spatially according to the corresponding regions in LiDAR and radar \acrshort{bev}s and then temporally across frames, following the two-stage spatial-temporal integration pattern observed in camera and LiDAR integration. 

There are few studies related to the temporal integration of camera, LiDAR, and radar combinations, possibly due to the complexity of the integration. Existing studies break down the multi-sensor temporal integration task in various ways. For example, in one study \cite{kim2020extended}, \acrfull{ekf} is deployed to each sensor respectively, and a reliability function is used for cross-modality integration. In another study \cite{shahian2019real}, the inputs of integration vary in different tasks, with camera images and LiDAR depth maps fused for 2D segmentation and object detection, while LiDAR and radar points over time are used for 3D obstacle detection and tracking. Liang \etal \cite{liang2020scalable} propose a loosely-coupled integration architecture for vehicle state estimation based on Error-state \acrshort{ekf}, where camera, LiDAR, and radar data are used as observations to correct the estimated priori state.

\section{Data Integration: When to Integrate}
\label{sec:when}


Since deep learning networks extract features with multiple neural layers, integration strategies can take place at different stages and in various ways \cite{feng2020deep}. Broadly speaking, the strategies are commonly classified into early (data-level), middle (feature-level), and late stages (decision-level). Data-level integration methods involve the fusion of raw or preprocessed data before feature extraction. Feature-level integration, on the other hand, combines extracted features at intermediate neural network layers. Decision-level integration involves the merging of output data separately estimated by each sensor. As features can be obtained at different depths of a neural network, several detailed integration paradigms have been designed and applied for feature-level integration. Feng \etal \cite{feng2020deep} further classifies the middle integration into three patterns, including fusion in one layer, deep fusion, and short-cut fusion. 

However, as described in Section~\ref{sec:intro-when}, this method fails to classify methods where integrated features are not at the same level (depth) of data abstraction, limiting its applicability to fit the increased amount of fusion schemes. Cui \etal \cite{cui2021deep} proposes an approach to incorporate more integration circumstances by introducing \quotate{multi-level} integration in addition to the original three classes. Thus, we review the related works with respect to these four categories in this section.

\subsection{Data-level integration}
Data-level integration is popular in SLAM \cite{li2018undeepvo, zhou2018deeptam, DBLP:conf/iros/JiaPCCLY20, wang2020tartanvo, li2019net, DBLP:conf/cvpr/ZhanGWLA018, fritsche2017modeling}, but rare in other tasks. \cite{zhou2018deeptam} uses optical flow to exploit relationship between frames to learn poses and uses cost volume between frames to learn depth maps. \cite{DBLP:conf/iros/JiaPCCLY20, wang2020tartanvo, li2019net} utilize convolutions to generate feature vectors. \cite{li2018undeepvo} stacks multi-frames together to regress pose. \cite{DBLP:conf/cvpr/ZhanGWLA018} regresses poses with convolutions. Fusing data in the early stage is also employed in detection \cite{moshiri2018pedestrian, ouyang2018multiview, li2019vehicle, kwon2016low}, depth completion \cite{fu2019lidar} and semantic segmentation \cite{wang2022lidarseg}. However, a large percent of these approaches lead to noncompetitive results.

\subsection{Feature-level integration}
Feature-level integration is widely adopted in most areas including object detection, depth completion \cite{yan2021rignet, cheng2019noise}, semantic segmentation \cite{zhou2022cross, meyer2019sensor, liu2020road, sun2020fuseseg, vachmanus2021multi, wei2022spatial, duerr2020lidar}, tracking \cite{shenoi2020jrmot, weng2020gnn3dmot, xu2019spatial, zhang2019robust, voigtlaender2019mots}, 3D reconstruction \cite{lei2022c2fnet, zhu2022garnet}, SLAM \cite{teed2021droid, wang2017deepvo, almalioglu2019selfvio, li2019sequential, zhang2022beverse, chen2023real}, action classification \cite{wei2022spatial, zhong2022no} and navigation \cite{cai2020probabilistic}. The popularity of feature-level integration comes from the extracted features, which are more compact and representative than the raw data. Nevertheless, feature-level merging has the adverse effect of diluting the single modalities’ strengths. The majority of 2D and 3D object detection methods fuse in feature level \cite{liang2018deep, wu2022sparse, bai2022transfusion, vora2020pointpainting, huang2020epnet, DBLP:conf/cvpr/BijelicGMKRDH20, fadadu2022multi, zhou2020end, Jianhui2018, meyer2019sensor, geng2020deep, schroder2019feature, mendez2021camera, nabati2021centerfusion, qian2021robust, wang2022detr3d, chen2022futr3d, liu2022petr, liu2022petrv2, yasuda2022multi, zhang2022beverse, liu2021multi, chen2022lane, yuan2021temporal, zeng2022lift, chen2022mppnet, kim2022camera, qi2022millimeter, zhang2022ri, li2023logonet, wang2023frustumformer, liu2022bevfusion}. 
\cite{liang2018deep} uses deep parametric continuous convolution layers with MLPs to directly output target feature in each \acrshort{bev} grid. \cite{li2023logonet} integrates LiDAR voxel features and image features with attention designs. 
Studies \cite{li2021hdmapnet, li2022bevformer, xiong2023neural, zhu2023nemo, liu2022petrv2, huang2022bevdet4d, zhang2022beverse} focusing on spatiotemporal fusion in \acrshort{bev} map generation integrate features of surrounding images or \acrshort{bev} feature maps at different timestamps with various designs of neural networks. 
In \cite{wu2022sparse}, features from both cloud points are grid-wisely combined based on learned weights. Besides, most 3D reconstruction methods \cite{DBLP:conf/cvpr/SunXCZB21, DBLP:conf/cvpr/LongLLTW21, DBLP:conf/eccv/YaoLLFQ18, DBLP:conf/cvpr/GuFZDTT20, DBLP:conf/iccv/MaGWHCY21, DBLP:conf/eccv/MurezABSBR20, DBLP:conf/cvpr/YangMAL20, DBLP:conf/cvpr/ChengXZLLRS20} integrate at feature level because of two reasons. On the one, the extracted features are more stable to environmental changes, e.g., lighting, which are crucial to photo-consistency measurement. Thus, it is less suitable to directly integrate raw data. On the other hand, the 3D estimates obtained from each single view is ambiguous in scale, which makes integrating at decision level challenging.

\subsection{Decision-level integration}
Decision-level integration has wide applications in many areas including semantic segmentation \cite{pan2020cross}, reconstruction \cite{tateno2017cnn, berrio2021camera}, object detection, tracking, SLAM and annotation \cite{wang2019latte}. Among all these domains, it is most popular in object detection \cite{obst2014multi, pang2020clocs, jha2019object, liu2020sensor, kowol2020yodar, wu2021pedestrian, dinesh2020stereo, PanHuihui2021, clunie2021development, clunie2021development}, tracking \cite{kim2021eagermot, cao2022observation, chiu2020probabilistic, weng20203d, osep2017combined, Sharma2018beyond, zhang2018vehicle, liu2021robust, ramos2017detecting, clunie2021development} and SLAM \cite{Czarnowski2020DeepFactors, DBLP:journals/jzusc/XueWDCHZ17, garcia2017sensor, liang2020scalable, kim2019fusing, kim2020extended, sun2022multi, tian2022accurate, zhang2022multiple}. Decisions from different sources can be either fused with learning-based strategies or non-learning-based ones.\cite{pan2020cross} adds up all input features from multiple views to generate fused features. \cite{tateno2017cnn} fuses the depth maps and uncertainty maps according to the weighted scheme, then output the tensor after weighted addition. \cite{obst2014multi} transforms cooperative awareness messages (CAM) to relative measurements, then uses buffering strategies to deal with time synchronization in out-of-sequence measurements (OOSM). \cite{pang2020clocs} fuses each pair of 2D bounding box (predicted from image) and 3D bounding box (projected to image) candidate by first generating a new tensor based on IOU and confidence, then further fuse with convolution layers.

\subsection{Multi-level integration}
Multi-level integration is adopted in lots of applications including object detection, depth competition \cite{hu2021penet}, semantic segmentation \cite{sobh2018end}, \cite{pfeuffer2019robust}, tracking, SLAM and 3D reconstruction \cite{DBLP:conf/cvpr/DuzcekerGVSDP21, lou2023slam}. Most multi-level integration methods are designed for object detection and tracking. There are also quite a percent of papers focusing on SLAM.
\cite{zhu2021vpfnet} fuses feature level information from the camera and data-level information from the LiDAR. \cite{paigwar2021frustum} and \cite{qi2018frustum} leverage the outputs of the image object detector as 2D proposals, which are then projected to form 3D searching space for 3D object detection. To specify, \cite{ravindran2022camera} projects LiDAR and radar information into 2D maps and use convolutions to fuse them with camera RGB. \cite{LIDAR_camera_fusion_for_road_detection_using_fully_convolutional_neural_networks} fuses features and raw data fusion via concatenation. \cite{dong2021radar} fuses decision level information from the camera and data level information from the radar. \cite{Dheekonda_2017} has one input at the object level and another at the feature level. \cite{hu2021penet} uses independent branches to extract features of images and LiDAR point cloud, and fuse (addition or concatenate) at multiple levels. \cite{DBLP:conf/cvpr/DuzcekerGVSDP21} warps the former depth map prediction to current view's hidden representation according to geometry. \cite{zhou2020tracking} uses tracking conditioned detector to detect objects in new frame (conditioned on previous frame and object detection result of previous frame) to get a temporally coherent set of detected objects, where the 2D displacement is predicted to associate detection results through time. \cite{yin2021center} uses multiple faces features to refine locations. \cite{DBLP:journals/ral/LuitenFL20} firstly uses optical flow to get short tracklets, then uses pixelwise depth estimation based on camera ego-motion to get 3D motion, finally uses 3D motion consistency to get long term tracklets. \cite{nguyen20203d} uses detected object position (decision-level) and feature vector from detected mask (feature-level) to associate frames, and uses nearby targets (neighbors) to constrain matching IoU distribution. \cite{frossard2018end} fuses 3D-2D detection by backproject 3D to 2D, and fuses frames by appearance association based on raw image and 2D/3D motion relation based on raw image and pointcloud. \cite{ravindran2022camera} fuses RGB features and raw data from LiDAR and radar. \cite{wang2020high} fuses the region predicted by camera and raw data from LiDAR and radar.

\section{Data Integration: How to Integrate}
\label{sec:how}

The ``how to integrate'' component of data integration involves the mathematical operations used to combine data/features. Feng \etal \cite{feng2020deep} summarize the fusion operations utilized in deep learning networks into four categories: addition/average mean, concatenation, ensemble, and mixture of experts. The characteristics of different integration operations are not further elaborated in their study. Wang \etal \cite{wang2023multi} categorize data combination approaches for 3D object detection into two primary types: alignment and fusion. Alignment is further subcategorized into projection-based and model-based methods, while fusion is divided into learning-based and learning-agnostic approaches. The authors of the present paper believe that alignment also serves as a means of fusion in many circumstances, and the model-based alignment and learning-based fusion techniques showcase significant overlaps with each other. Besides, since both \cite{feng2020deep} and \cite{wang2023multi} focus on methodologies for specific perception tasks (object detection and semantic segmentation), other integration operations such as \acrshort{ekf} \cite{dinesh2020stereo, osep2017combined} or probabilistic map \cite{ramos2017detecting, berrio2021camera} are not discussed.

In Table~\ref{tab:operations}, we provide a brief summary of several commonly used operations in deep learning-based data integration processes, including projection, concatenation, addition-similar operations (addition/weighted summation/average mean), probabilistic methods, rule-based transaction, temporal integration approaches, and encoder-decoder methods. The application of integration operations may be highly dependent on the task, inputs, and data abstraction levels. Some operations are more versatile and applicable in various conditions, while others are more specific and implemented for certain purposes. Additionally, an integration algorithm may involve multiple different operations at various stages.

\subsubsection{Projection}
\label{sec:how-projection}

Projections, such as 3D-to-2D projection and 2D-to-3D back-projection (also known as ``reverse projection''), are commonly used to connect images in 2D space and point clouds in 3D space, allowing for data to be operated within the same domain. 3D-to-2D projection reduces the dimensionality of 3D objects to a 2D plane, while 2D-to-3D back-projection may rely on geometric cues or depth information. Projections are frequently utilized to generate different data representations \cite{wang2023multi}. For example, pixel-based view representations of LiDAR points and point-based pseudo-LiDAR point representation of images, as mentioned in Section~\ref{sec:sensor-lidar}, are outputs of projection and back-projection, respectively. Additionally, projection can also serve as an integration operation to combine data. 

As an integration operation, projection may take place at raw-data level, feature level, or decision level. 
\begin{itemize}
	\item Raw-data level integration refers to projecting LiDAR or radar point clouds onto 2D plane such that they can be combined with corresponding RGB image information \cite{wu2022sparse, berrio2021camera, dong2021radar, wang2019latte, vora2020pointpainting, meyer2019sensor, fei2020semanticvoxels}. An example is LATTE \cite{wang2019latte}, where the points are projected onto the RGB image plane at the beginning of the sensor fusion pipeline. Similarly, LiDAR points are projected onto the image to obtain segmentation information in PointPainting \cite{vora2020pointpainting}. In \cite{lou2023slam}, dense 3D points depth map is generated via projecting 3D LiDAR points to the corresponding semantic 2D image followed by interpolation. 
	\item Feature-level projection is to project features extracted from LiDAR or radar points to 2D space for integration. This integration usually takes place in the middle layers of a neural network. An example is EPNet \cite{huang2020epnet}, where point features and image semantic features are combined in multiple scales in 2D domain. In LoGoNet's global and local fusion modules \cite{li2023logonet}, voxel point centroids or center points are projected to the image plane to generate reference points to sample image features, and utilizes attention structure for integration. A problem for raw-data and feature level projections is how to deal with the resolution consistency between LiDAR and camera branches. In \cite{meyer2019sensor}, point clouds are projected onto extracted RGB feature maps which have lower resolution to avoid discarding image information.  
	\item Decision-level 3D outcomes are projected to 2D space for integration in some other works. For example, in \cite{kim2021eagermot}, 3D bounding boxes obtained from point clouds are projected to 2D image plane, such that the original 3D bounding boxes and 2D bounding boxes generated from image can be associated based on box overlap in image domain. \cite{kowol2020yodar} projects radar points to image plane, and integrates radar predictions (slices in this paper) and camera predictions based on box overlaps. In the early stage of CAMO-MOT \cite{wang2023camo}, 3D detection results of LiDAR point cloud are projected onto the pixel plane of the camera to obtain 2D detection results.  
\end{itemize}

A few works conduct decision level back-projection to integrate 2D image data with 3D points. A common operation for object detection is to back-project 2D proposals to 3D space, generating frustums as the regions of interest (RoI) to guide 3D point searching and crop clouds \cite{qi2018frustum, paigwar2021frustum, wang2020high, wang2020high}. To reduce the irrelevant background information wrapped in RoIs, Yang \etal back-projects segmentation masks instead of bounding boxes to the point clouds \cite{yang2018ipod}. Since LiDAR points capture geometric information and by nature can easily be clustered to distinguish foreground from background, the points belong to background can be projected back onto the image to refine 2D segmentation. This types of integration operation based on back-projection is called ``ensemble'' \cite{feng2020deep}. However, since this back-projection operation integrates LiDAR points and images in a data association or supervision way, it usually fails to fully exploit the information in images and is considered as ``weak fusion'' in \cite{huang2022multi}. 

After reviewing the aforementioned methods, we identify two common issues that need to be addressed when integrating data through projection techniques.
For the first, since multi-modality data are collected with different sampling rates and in different coordinates, they must be carefully aligned into a unified coordinate before projection and mapping. This process heavily relies on the sensor extrinsic and intrinsic matrices, which may be vulnerable to parameter errors. For the second, how to deal with the resolution inconsistency may significantly impact the performance of the integration \cite{meyer2019sensor}. In existing works, these two issues are usually not fully explored or discussed.

\subsubsection{Concatenation}
\label{sec:how-concatenation}

Concatenation is an operation widely used to combine data or feature maps at different layers of a neural network \cite{zhu2021vpfnet, hu2021penet, yan2021rignet, jha2019object, yu2019multi, cai2020probabilistic, sun2020fuseseg, oliveira2020topometric, el2019rgb, farahnakian2020rgb, xiao2020multimodal, li2022deepfusion, lei2022c2fnet, kim2022camera}. Concatenation operation can either be in the way of stacking feature maps along the depth as additional channels, or be added to the end of flattened vectors. 
For example, \cite{ouyang2018multiview} and \cite{shahian2019real} convert LiDAR point clouds to gray images, and concatenate with RGB image as additional channels. Similarly, the projected and processed LiDAR and radar channels are concatenated with camera images for multiple times in \cite{ravindran2022camera}. In \cite{weng2020gnn3dmot}, the image feature vector is obtained by concatenating motion and appearance feature vectors obtained from ResNet \cite{he2016deep}. Then the image and LiDAR features are integrated again by concatenation. 
Qi \etal \cite{qi2022millimeter} combine camera and radar information by concatenating two-channel radar image with same-sized 64-channel visual image feature maps. 
MVFusion \cite{wu2023mvfusion} designs a semantic-aligned radar encoder (SARE) in which semantic indicator produced via all stages' visual features are concatenated with radar inputs. 
In C2FNet's \cite{lei2022c2fnet} fine generation module, concatenation takes place to integrate global guidance features with point cloud features in different levels and in different views. In GARNet \cite{zhu2022garnet}, the feature map of each view is concatenated with its deviation from the global feature maps in post merger block. One study \cite{caltagirone2019lidar} presents trials of early fusion approach in which LiDAR and camera data can be concatenated in depth dimension, and late fusion approach in which the multi-modality features are concatenated. Besides, concatenation can be used as a straightforward temporal fusion approach. In \cite{huang2022bevdet4d, liu2022petrv2}, \acrshort{bev} features of previous frame are spatially aligned and concatenated with the ones of the current frame. 

Concatenation in depth dimension requires the inputs to be in the same spatial size. Specifically, the width and height of feature maps to be concatenated together should be the same, while the number of channels can be different. For example, in MVFusion \cite{wu2023mvfusion}, raw radar points are preprocessed into a representation with the same shape as images before concatenation, by first extending them to pillars and then projecting pillars to the corresponding image view.
Because of this, concatenation can be used to combine data from more than two data sources. In \cite{cai2020probabilistic}, information from camera, LiDAR, radar, and localization and mapping are concatenated together in perception module. In \cite{gao2023spatio}, input features pairs for cross-frame aggragation are the concatenation outcomes of template feature, reference frame feature, and reference frame point. 

Concatenation is a straightforward operation with numerous applications, particularly as a fundamental combination technique in neural networks. The output format of concatenation closely resembles the input format, although the dimension might alter (for example, an increase in the number of channels or vector length). Nonetheless, concatenation has certain limitations in practice, apart from the stringent requirement of equal spatial size. Firstly, the input dimension of concatenation is typically inflexible. Consider a camera-LiDAR integration model where 3-channel LiDAR data is combined with 3-channel RGB data (3 RGB channels initially, followed by 3 LiDAR channels). If one sensor fails, the model will not function correctly because the integrated data's dimension decreases from 6-channel to 3-channel, making it incompatible with downstream modules. Secondly, concatenation input sensors are not interchangeable. If the order of the two sensors is reversed, the model will not produce accurate results. As a result, concatenation struggles with varying input dimensions and is not ideal for integration systems that utilize interchangeable sensors.

\subsubsection{Addition-similar operations}
\label{sec:how-addition}

Addition, weighted sum, and average mean are mathematical operations used to combine two data sources by adding them together with pre-determined weights. These operations can be applied to values, feature vectors, or feature maps in an element-wise way. An example of integration with addition can be found in \cite{pan2020cross}, where the authors add all features up in the multi-view fusion module. Element-wise summation is also implemented in \cite{liang2019multi, lv2022maffnet}. As shown in the residual fusion module in \cite{yu2019multi}, multiple feature maps from LiDAR and camera are integrated together via element-wise mean. One disadvantage of addition and average mean is that they combine different data sources with equal importance. Weighted sum makes it possible to give different weights to different inputs. In the depth map refinement stage of the model in \cite{tateno2017cnn}, the depth maps and a uncertainty map are fused according to a weighted scheme. In \cite{wu2022sparse}, the authors predict a pair of weights for each pair of the LiDAR and image features, thus the features can be integrated considering the reliability of the information. In \cite{zhu2022garnet}, score maps are generated in merger blocks as weights when fusing feature maps and when fusing the reconstructed voxels from all view images.

Addition-similar operations are one of the simplest way for integrating features. While these operations are simple to implement, it is required that the two parties to be added to have the same spatial size and depth, \ie to be in the exactly same format. The format of the output after these operations is also the same as the format of each input side. This is an even harder constraint comparing with concatenation: two feature maps can be concatenated if they can be added, otherwise not. This enables some researchers to compare addition and concatenation in their integration models \cite{hu2021penet, weng2020gnn3dmot, zhang2019robust}. Another disadvantage of addition and average mean is that both parties to be combined have same weights, no matter how confident the respective data are. This may cause problems when one sensor fails to work or has low data quality. Also, as raised in \cite{feng2020deep}, while addition and average mean help to achieve high average precision, the network may fail in corner cases thus have low robustness.

\subsubsection{Probabilistic methods}
\label{sec:how-probabilistic}

Probabilistic integration methods incorporate uncertainties into the integration process in different ways. Berrio \etal \cite{berrio2021camera} assign a vector to represent probabilities of different semantic classes to each LiDAR point when doing back-projecting. Another segmentation research \cite{park2021drivable} also uses a probabilistic method to help select the optimal segmentation result from independent outputs of image and LiDAR. \cite{ramos2017detecting} and \cite{wu2021pedestrian} utilize Bayesian rules to obtain the confidence of object detection after integration. \cite{garcia2017sensor, liu2021robust, shenoi2020jrmot} leverage Joint Probabilistic Data Association (JPDA) filter based on \acrfull{kf} methods for system matching and state updating in tracking. Different from \acrshort{kf} methods that are usually used to update the status of a single track given an observation, JPDA takes all measurements and tracks in the scene into account with joint probabilities in order to obtain a better estimation. 

Incorporating uncertainties in the integration pipeline is essential for autonomous driving applications. This mechanism helps the system to assess risks and adjust confidences of choosing trustworthy sensory information for decision making. With this, the integration system can be more robust to different external environment and varying sensor data quality. However, they also have more parameters to be estimated and thus are usually applied to comparatively simple models.

\subsubsection{Rule-based transaction}
\label{sec:how-rulebased}

There are a few researchers adopt handcrafted rules to guide the integration. \cite{pang2020clocs} encodes each detected 2D and 3D bounding box pairs into a 4-channel tensor, while the 4 elements of the tensor are chosen and designated carefully. \cite{fritsche2017modeling} presents a method to integrate radar and LiDAR measurements/features by replacing low-quality LiDAR points affected by fog with radar data. To achieve this, the authors propose a set of manually-designed heuristic rules according to different conditions. Wang \etal \cite{wang2020high} generate a 7-dimensional frustum by integrating RGB and LiDAR, and 8-dimensional points based on manually selected metrics of radar point clouds. Though these integration operations are proved to be effective in these papers, they may hardly be implemented in another scenario. Comparing with other integration operations, aforementioned rule-based ones are dedicatedly designed for one task or dataset, thus have poor generalizability.

\subsubsection{Temporal integration approaches}
Non-deep-learning approaches for temporal information integration such as \acrshort{kf} and relevant extensions (e.g., \acrshort{ekf}, Unscented Kalman Filter) have long been the mainstream methods for multi-frame data association in localization and tracking \cite{kim2021eagermot, cao2022observation, wang2022lidar, chiu2020probabilistic, weng20203d, zhang2018vehicle, liu2021robust, gohring2011radar, osep2017combined, garcia2017sensor, liang2020scalable, shahian2019real, kim2020extended, zhang2022multiple, camarda2022multi}. In general, the pattern of tracking with \acrshort{kf} methods is to first utilize feature extractor or detector to get the information of the object to track, and then employ \acrshort{kf} models to integrate frames.

Deep learning approaches with comparatively intuitive and interpretable structure, such as RNN family (e.g., RNN, LSTM/ConvLSTM, GRU/ConvGRU) and 3D CNN, are designed to leverage temporal information \cite{griffin2021depth, patil2020don, zhang2019exploiting, liu2020multi, cs2018depthnet, qian2021robust, duerr2020lidar}. Similar to \acrshort{kf}, the pattern of temporal integration with RNN models also has two stages: first single-frame object detection with various methods, and then temporal integration with RNN family models. ODMD proposed by Griffin \etal \cite{griffin2021depth} utilizes 2D object bounding boxes of a target in two frames and corresponding camera motion to estimate the target's depth with LSTM. \cite{liu2020multi, patil2020don, cs2018depthnet} integrate information of previous frames and generate depth maps with ConvLSTM.
DeepVO uses RNN to fuse extracted features \cite{wang2017deepvo}, and \cite{li2019sequential, xue2019beyond} leverage LSTM to integrate optical flows or feature maps in time sequence. Besides, a deep learning Visual Inertial Odometry (VIO) method uses CNN and LSTM to get image feature map and \newacronym{imu} features respectively, and then conducts temporal integration with LSTM \cite{clark2017vinet}. 
Object detection and \acrshort{bev} map generation in \cite{liu2020understanding} employ LSTM to integrate temporal information. In \cite{xiong2023neural, zhu2023nemo}, \acrshort{bev} map grids are also temporally fused and updated with recurrent network designs. 3D CNN method is extended from traditional 2D CNN by stacking 2D feature maps at different timestamps together and utilizing a 3D convolution kernel to extract and integrate spatio-temporal information. Different from RNN models where the spatial and temporal feature extractions are conducted in two independent stages, 3D CNN kernels can extract and exchange spatial and temporal information simultaneously. This structure has been used for video image processing, recognition, and analysis in many different realms \cite{diba2018spatio, chen2019med3d, hara2018can, feichtenhofer2020x3d}. In autonomous driving area, Qian \etal \cite{qian2021robust} propose MVD-Net, a 3D CNN network combining LiDAR and radar signals, to detect vehicles via spatio-temporal integration. In BEVerse \cite{zhang2022beverse}, \acrshort{bev} features generated at consecutive timestamps are warpped and aligned to form a stack of temporal block, and fused with 3D convolutions. 

\acrshort{kf} and its variants, RNN families, and 3D CNN are commonly employed specifically for temporal data association and integration, yet they have their own limitations. \acrshort{kf} methods heavily rely on the manually designed process model and the estimation of covariance matrix, which may not be appropriate or accurate. RNN methods tend to separate spatial and temporal integration into two independent stages. Though this order brings conveniences for model tuning, whether it is the optimal spatial-temporal integration paradigm still remains in doubt. Another disadvantage of RNN family models is that they usually have poor capability to track the long-term dependency and are difficult to train and converge. 3D CNN models integrate spatial-temporal information at the same time. However, dense 3D convolution networks usually have large number of parameters to be estimated, which may easily result in over-fitting when the training set is not large enough, or limited spatial and temporal reception fields if the kernel size is small.  

\subsubsection{Encoder-decoder methods}
\label{sec:how-nn}

Many researchers exploit neural networks or encoder-decoder architecture models to integrate multi-view, multi-modality, and multi-frame sensory data. CNN network and its derivations are the most common neural networks used for feature extraction and integration. These methods are data driven, which means that they do not rely on handcraft-designed models. The kernels and convolution layers by nature can integrate spatial information of the input data. In many cases, these networks utilize well-designed combinations and procedure of multiple aforementioned integration operations and convolution layers \cite{liang2019multi}. In \cite{zhao2018object}, the authors first use different sub-networks to extract features from image and proposal regions from multiple LiDAR view-representations and the original 3D point cloud. Then the proposals are projected into image tensors at multiple layers of their neural network. Caltagirone \etal \cite{caltagirone2019lidar} propose an innovative cross-fusion network architecture, in which the two input branches (2D RGB and LiDAR images) are connected by trainable scalar cross connections. Specifically, at each layer, the input feature tensors of LiDAR and camera are added together with learnable weights. \cite{ravindran2022camera} introduces Bayesian Neural Network along with concatenation operations in each layer to capture the uncertainties with the cost of almost doubled number of parameters to be estimated. 

Though CNN models perform well in their respective fields, there are limitations for data integration. Firstly, they are also subject to the shortcomings of the integration operations included in the model structure. That is to say, for instance, a CNN model with concatenation operation in the network still suffers from the weakness of concatenation. Secondly, the mechanism of CNN and their limitations in the receptive field hinder its ability to perceive and integrate global information. Furthermore, the model procedure and structure are unchanged in application, which means that the integration is not adaptive to the changing environment.

Besides CNNs, an increasing amount of researchers set their sights on attention-based models (e.g., Transformer) for feature combination. Chen \etal \cite{chen2022futr3d} integrates camera, LiDAR, and radar data with Transformer for 3D object detection. A set of queries encoding 3D locations can be projected to the corresponding input space of each sensor and get their features.  
DETR3D \cite{wang2022detr3d} proposes a 3D object detector utilizing Transformer to integrate multi-view image information. LoGoNet \cite{li2023logonet} integrate multi-modality data with attention blocks by setting LiDAR features as \textbf{Q} (query) and image features as \textbf{K} (key) and \textbf{V} (value). 
In the radar-guided fusion transformer (RGFT) module of MVFusion \cite{wu2023mvfusion}, cross-attention mechanism is introduced to fuse radar and image features. 
\cite{bai2022transfusion} leverages two Transformer decoder layers to produce 3D bounding boxes with queries that associate both LiDAR BEV features and image features. 
BEVSegFormer adopts \acrshort{bev} queries to combine multi-camera features in cross-attention module \cite{peng2023bevsegformer} to obtain \acrshort{bev} segmentation results. 
RI-Fusion \cite{zhang2022ri} presents an attention module that merges range view feature derived from LiDAR data and RGB image feature. Specifically, the former is transformed into \textbf{Q} while the latter is converted to \textbf{K} and \textbf{V}. FrustumFormer \cite{wang2023frustumformer} designs both scene queries and instance queries to transform multi-scale multi-view image features to a unified \acrshort{bev} feature with cross-attention layers. 

In addition, recent trials in various domains start focusing on Transformer to integrate spatial and temporal information \cite{zhou2022transvod, bertasius2021space, li2022uniformer, li2022bevformer, zeng2022lift, jiang2022polar, pang2023standing}. Theoretically, Transformer have unlimited receptive fields and can obtain global information with the \textbf{Q}, \textbf{K}, and \textbf{V} structure. BEVFormer \cite{li2022bevformer} integrate camera image multi-view and multi-frame information with Transformer architecture. Decoder of BEVFormer has a temporal self-attention layer where BEV queries can interact with historical BEV features, and a spatial cross-attention layer where BEV queries interact with features of other camera views. \cite{zeng2022lift} is another example leveraging Transformer to integrate spatial-temporal information. The authors process camera and LiDAR features with cross-sensor point-wise attention, and integrate grid-wise features from BEV maps obtained at different timestamps as 4D tensors. \cite{sun2020transtrack, zeng2021motr, meinhardt2022trackformer} propose different designs of cross-frame query-key structures for multi-object tracking over time given image sequences as inputs. Study \cite{wang2023frustumformer} utilize deformable cross-attention to aggregate information of history queries into current instance queries.
Comparing with CNN, Transformer can has significantly larger, or global, receptive field. Its flexible model design also makes it possible to applied to a wide range of applications. Furthermore, Transformer is also scalable to model any data size. Though models utilizing Transformer structure tend to be larger and harder to train, its advantages of high structure design freedom and cross-domain information interaction architecture make it very suitable for data integration. We believe that more diverse Transformer-based spatial-temporal integration designs will spring up in the near future.

\section{Case Study}
\label{sec:casestudy}

\begin{table*}[!h]
	\small
	\centering
	\begin{tabular}{cccccccccc}
		\hline
		Camera & Number of sensors & Working range (meters) & Applicable scenarios\\
		\hline
		Rearview & 1 & 50 & Objects in the back \\
		Wide forward & 1 & 60 & Traffic lights, short-distance or crossing objects \\
		Forward-looking side & 2 & 80 & Neighboring or crossing objects\\
		Rearward-looking side & 2 & 100 & Objects in the neighboring lanes behind the car\\
		Main forward & 1 & 150 & Most cases\\
		Narrow forward & 1 & 250 & Distant objects\\
		\hline
	\end{tabular}\\
	\caption{The cameras adopted by Tesla.}
	\label{tab:tesla cameras}
\end{table*}

Due to the prosperity and rapid development of autonomous driving technology in recent years, many companies, such as Tesla, Waymo, and General Motors, have launched \acrshort{ads}-related products and services via different technical paths. In this section, we take Tesla's vision structure as an example to illustrate data integration applications in real-world \acrshort{ads} systems with the \quotate{what-when-how} analysis structure we described previously. 

Tesla builds a pure camera-based system, which consists of eight cameras in total. Table~\ref{tab:tesla cameras} displays the perception range of the mounted cameras. These cameras have the field of view overlapped with each other and together cover 360\textdegree\ around the car. The three cameras (main/narrow/wide forward) installed behind the windshield help to detect distant objects within a broad view angle. Specifically, the wide camera, equipping a 120-degree fish-eye lens, is helpful in downtown and crossroads where vehicles drive slow. The narrow camera is applicable on the highway. Besides the front sensors, there are also cameras observing both sides and the rear of vehicles. The forward-looking side cameras with a view angle of 90 degrees are mainly used to detect cars that change lanes and objects around crossroads. The rearward-looking side cameras assist lane changing. The rear-view camera avoids rear-end collisions and facilitates parking.

All the cameras have different directions and are capturing images at certain frequencies, so that the integration of Tesla's autonomous driving system can be regarded as multi-view and multi-frame integration.

Similar to the approach we described earlier, Tesla splits multi-view multi-frame integration into two steps: first, multi-view spatial integration, and then multi-frame temporal integration. Integration between different views and frames is done at the feature level. After raw image calibration and multi-scale feature extraction, a transformer-like neural network is used to fuse features from different views. Specifically, the key and value of the transformer come from the features of the images, while the query comes from the position encoding of each point in the occupancy space. The occupancy space mentioned here is similar to the \acrshort{bev} space, which is a top-down view, but with an additional dimension of height. Each point in the occupancy space queries its most relevant corresponding object in other views through the attention mechanism. After aggregating the spatial information, the model aligns the positions of objects at different times using displacement information provided by the odometer for temporal fusion, with features from nearby time steps having higher weights. The spatiotemporal features are then passed through a set of deconvolutions to obtain high-resolution 3D features in the occupancy space for further decision-making.

\section{Conclusion and Discussion}
\label{sec:conclusion}

In this paper, we review the latest deep learning-based data integration techniques for autonomous driving perception, focusing on three types of sensors: cameras, LiDAR, and radar. The need for data integration arises from the complementary nature of perception capabilities among these sensors. We present a unique perspective on integration techniques by examining ``what, when, and how to integrate.''
Under ``what to integrate,'' we propose a novel taxonomy that classifies data integration inputs into seven categories based on three dimensions (multi-view, multi-modality, and multi-frame). We observe that most existing research focuses on one-dimensional integration, while two-dimensional integration is slowly gaining interest. For ``when to integrate,'' we categorize data integration techniques based on data abstraction levels. In ``how to integrate,'' we not only discuss commonly used data/feature combination operations but also comprehensively outline their advantages and drawbacks, which is seldom addressed in previous studies. Furthermore, we demonstrate data integration techniques using Tesla's \acrshort{ads} as an example.

By examining related works published within the past five years, we identify several patterns or issues in current integration methodologies:

1) Spatial alignment based on explicit projection: As discussed in Section~\ref{sec:how-projection}, projection and back-projection are two commonly used integration methods for spatially aligning data from different sensors. These methods rely on the geometric relationship between sensory measurements and intrinsic and extrinsic parameters for mutual projection. However, due to dimension and perspective differences between sensory measurements and estimation errors in the transformation matrix, these approaches can lead to information loss, redundancy, or cumulative error.


2) High specificity but low generalizability: Many existing data integration methods/algorithms are designed for specific scenarios, making them sensor-specific, data-representation-specific, and task-specific. While this high specificity contributes to improved perception performance in their respective domains, it also results in low generalizability, limiting the applicability of each integration technique.


3) Fixed integration architecture with limited flexibility: Most existing integration techniques have fixed structures, which means that the weights, orders, dimensions, formats, and depths of the integration models usually do not change once a neural network is trained and ready for implementation. This property makes the model unable to handle some common practical situations, such as unavailability of data from one or several sensors (not sensor pluggable) or changes in the order of the input (not input permutation invariant). The unchangeable network is also difficult to scale from simple to complex applications, for example, extending from two-frame integration to multiple consecutive frames. Moreover, these integration architectures cannot automatically adapt to external changes and select the best integration method by adjusting their structure and weights.

Comprehensive understanding of the aforementioned shortcomings sheds light on required properties of ``ideal'' approach for data integration:

1) Task- and modality-agnostic: Developing high-specificity individual integration methods for every scenario and task can result in redundant computation. Therefore, integration models that can be generalized and applied to a wide range of application scenarios is a future direction of development. Though existing research have provided much experience of integration, their nature of task-, representation-, and modality-specificity may also bring limitations when generalizing to other applications.

Developing Task- and modality-agnostic integration approaches, referring to data integration techniques without assumptions based on existing heuristic knowledge of sensors or different perception tasks, are essential for multi-view, multi-modality, and multi-frame integration. Specifically, in task- and modality-agnostic integration, the input data representations may not be described by one type of sensors or another, and the integration method may not be determined by one task. 

2) Sensor pluggable, permutation invariant, and spatial-temporal scalability: In practical applications, information missing may happen to a sensor for various reasons. Camera may not be able to take pictures if its lens is temporarily covered or contaminated. However, the perception module of the autonomous driving system can not fail due to this. An ideal integration approach is expected to provide comparatively high-quality integration outcomes even when a portion of information is lost. This also means that the data integration approach should be able to process data coming from different sensor configurations. In other words, the sensors are pluggable for the integration approach. 

To be able to process data coming from different sensor combinations, the integration approach needs to be invariant to input permutation, which means that the integration and perception performance should not be affected by the order of the inputs (e.g., not affected by whether LiDAR data or image data arrives first). 

Spatial-temporal scalability describes the capability of the integration method to adapt to increased demand both in space and time. Specifically, this refers to the ability of integrating a larger amount of data from more sensors, and integrating temporal information from more frames in a wider time window. This capability enables self-driving vehicles to deal with complex driving scenarios, in which the demand for processing spatial-temporal data is high.

3) Adaptive: Traffic conditions are changing all the time with their own levels of complexity. For example, car following on highways with comparatively simple traffic scenes and limited road elements can be processed without complicated data integration, while driving through complex and crowded intersections in cities may require much more information to process for decision making. Handling perception demands of different scenes by integration data in the same way may lead to unnecessary computational consumption or errors due to under-computation. Thus, another important property of dynamic integration approach is adaptiveness. 

The ideal integration approach is expected to adaptively learn the demands of perception according to the complexity of the scene, and thus adjust its structure, \ie activating different neurons, to fit the requirement of different driving environments. In low-complexity conditions where part of sensory measurements is not needed, these sensors can be ``unplugged'' from the integration processing (not integrated for simplicity). In temporal integration, the ``ideal'' integration approach can also select the appropriate fusion window or key frame according to the importance of the previous frames.

\section*{Acknowledgments}
The authors would like to thank Jiaheng Yang from Riemann Laboratory, Huawei Technologies, for his inspiring discussions. The authors also thank Qirui Wang from Parallel Distributed Computing Laboratory, Huawei Technologies, for collecting literature.

\bibliographystyle{IEEEtran}
\bibliography{ref}

\newpage
\onecolumn

\begin{landscape}
\appendices
\section{}
\label{appendix:A}

\begin{table}[h]
\caption{Summary of Recent Reviews on Data Integration Techniques in ADS Perception with Deep Learning Approach. \label{tab:reviews}}
\small
\centering
\begin{tabular}{p{\dimexpr0.09\linewidth-2\tabcolsep-\arrayrulewidth\relax}
                p{\dimexpr0.13\linewidth-2\tabcolsep-\arrayrulewidth\relax}
                p{\dimexpr0.2\linewidth-2\tabcolsep-\arrayrulewidth\relax}
                p{\dimexpr0.25\linewidth-2\tabcolsep-\arrayrulewidth\relax}
                p{\dimexpr0.33\linewidth-2\tabcolsep-\arrayrulewidth\relax} 
              }
\hline
Reference & Sensors & Applications/Tasks & Fusion Taxonomy & Other Contents\\
\hline 
Wang et al. \cite{wang2019multi} & Camera, LiDAR, MMW-radar, GPS, IMU, ultrasonic, V2X & Multi-target tracking and environment reconstruction (motion model and data association) & Discernible units (data level), feature complementarity (feature level), target attributes, decision making (decision level) & Present characteristics, advantages, and disadvantages of different sensors.\\

Fayyad et al. \cite{fayyad2020deep} & Camera, LiDAR, MMW-radar, GPS, IMU, INS, map & Detection, ego-localization and mapping & Data level (early fusion), feature level (mid-level fusion), or decision level (late fusion) & A summary of deep learning algorithm architectures in the field of sensor fusion for autonomous vehicle systems.\\

Cui et al. \cite{cui2021deep} & Camera, LiDAR & Depth completion, object detection, semantic segmentation, tracking and online cross-sensor calibration & Signal level, feature level, result level, and multi-level & A summary of deep learning fusion methods based on different sensor combinations and different input representations for each perception task. \\

Yeong et al. \cite{yeong2021sensor} & Camera, LiDAR, radar & Obstacle detection & Low-, mid-, high-level fusion & 1. Operating principles and characteristics of sensors, and a comparison of commercially hardware. \newline 2. Sensor calibration overview. \newline 3. A summary of related data fusion reviews.\newline 4. A summary of recent studies on sensor fusion technologies. \\

Huang et al. \cite{huang2022multi} & Camera, LiDAR & BEV object detection, 3D object detection & New taxonomy of multi-modal fusion: two major classes (strong- and weak-fusion), and four minor classes in strong-fusion (early-, deep-, late-, and asymmetry-fusion) & 1. Summary of data formats and representations of LiDAR and camera. \newline 2. Summary of commonly used open datasets.\\

Feng et al. \cite{feng2020deep} & Camera, LiDAR, radar & Object detection, semantic segmentation & Early-, late-, middle-fusion (fusion in one layer, deep fusion, short-cut fusion) & 1. Discusses the fusion methodologies regarding ``what to fuse'', ``when to fuse'' and ``how to fuse''. \newline 2. Summary of multi-modal datasets and task-related algorithms in papers.  \\
 
Wang et al. \cite{wang2021multi} & Camera, LiDAR & 3D object detection &  Feature fusion (granularity: RoI-wise, voxel-wise, point-wise, pixel-wise), decision fusion. & 1. Summary of popular sensors in ADS, their data representations, and corresponding object detection deep learning networks. \newline 2. Datasets (and metrics) for multi-modal 3D object detection. \\

Wang et al. \cite{wang2023multi} & Camera, LiDAR, radar & 3D object detection & New taxonomy with aspects of representation, alignment, and fusion. Fusion methods are further divided into learning-agnostic based (element-wise operations and concatenation) and learning-based (attention mochanism) approaches. & 1. Categorization of sensor data representation: unified representation (hybrid-based, stereoscopic-based, and \acrshort{bev}-based) and raw representation. \newline 2. Categorization of alignment methodology: projection-based (global projection and local projection) and model-based (cross-attention).  \newline 3. Datasets (and metrics) for multi-modal 3D object detection. \\
\hline
\end{tabular}
\end{table}

\section{}
\label{appendix:B}

\begin{table}[h]
\caption{Summary of Integration Operations in ADS Perception with Deep Learning Approach. \label{tab:operations}}
\small
\centering
\begin{tabular}{p{\dimexpr0.2\linewidth-2\tabcolsep-\arrayrulewidth\relax}
                p{\dimexpr0.3\linewidth-2\tabcolsep-\arrayrulewidth\relax}
                p{\dimexpr0.3\linewidth-2\tabcolsep-\arrayrulewidth\relax}
                p{\dimexpr0.2\linewidth-2\tabcolsep-\arrayrulewidth\relax}
              }
\hline
Operations / Methods & Advantages & Disadvantages  & Output Format\\
\hline
Projection & 1. Can connect 2D space and 3D space. 2. Based on the projection principle, which is easier to understand.  & 1. Heavily rely on the sensor extrinsic and intrinsic matrices, which may be vulnerable to errors. 2. Need extra methods to deal with revolution inconsistency before projection.  & Projected data with dimension reduction or dimension boosting. \\

Concatenation & 1. Easy to operate. 2. Can integrate data from more than two sources. & 1. Can only applied to data with same spatial size. 2. Can not deal with missing dimensions. 3. The input sensors are not permutable.  & Concategated tensors with same spatial size but different dimension. \\

Addition / Average mean \newline / Weighted sum & Simple operation and easy to implement. Wide range of applications at different stages of the integration process. & Require the data to be integrated to have exactly the same format (both spatial size and dimension). & Added data with same data format (both spatial size and dimension) as the input. \\

Probabilistic methods & Incorporate uncertainties into the integration process. & Usually have more parameters to be estimated thus currently are applied to comparatively simple models. & Depend on the input and implementation scenes. \\

Rule-based transaction \newline or integration & High specificity to deal with a certain scenario or dataset. & Low generalizability, and usually cannot be applied to other scenes. & Depend on the input and implementation scenes. \\

Kalman Filter and \newline extensions & 1. Easy to apply to integrate temporal information. 2. Do not acquire massive data to process and update. Can update system state with only a few data points. & 1. Heavily rely on the manually designed process model and the covariance matrix, which may not be appropriate or accurate. 2. Have short memory for temporal information. & Depend on the design of the system states. \\

RNN family & 1. Designed for temporal integration by nature. 2. Comparatively easy to understand and implement. & 1. Have poor long-term memory. 2. Difficult to train and converge.  & Concatenated tensors. \\

CNN, 3D CNN, and variants & 1. Data driven, and does not need manually designed system model. 2. Kernels and convolution layers by nature can integrate spatial (CNN) and temporal (3D CNN) information.  & 1. Comparatively limited receptive field and memory. 2. Subject to the shortcomings of the integration operations included in the model structure. 3. Model procedure and structure are unchanged in application. & Concatenated tensors. \\

Transformer & 1. Can potentially integrate both spatial and temporal information. 2. Global receptive field. 3. Flexible model design and scalable to model any data size. & 1. Need well-designed Queries and positional encodings. 2. Hard to train. & Tensor weighted summation. \\

\hline
\end{tabular}
\end{table}

\end{landscape}

\vfill

\end{document}